\documentclass{article}

 \usepackage[main, final]{neurips_2026}

\usepackage[utf8]{inputenc} 
\usepackage[T1]{fontenc}    
\usepackage[colorlinks=true]{hyperref}       
\usepackage{url}            
\usepackage{booktabs}       
\usepackage{amsfonts}       
\usepackage{nicefrac}       
\usepackage{diagbox}
\usepackage{microtype}      
\usepackage{xcolor}         
\usepackage[most]{tcolorbox}
\usepackage{algorithmic}
\usepackage{algorithm}
\usepackage{graphicx}
\usepackage{enumitem}
\usepackage{wrapfig}
\tcbuselibrary{skins,breakable}
\usepackage{caption}
\newtcolorbox{ruledalgobox}{
  enhanced,
  breakable,
  colback=white,
  colframe=black,
  boxrule=0.6pt,
  arc=1.2pt,
  left=6pt,right=6pt,top=6pt,bottom=6pt,
  fonttitle=\bfseries,
  title style={draw=none, fill=none},
}

\usepackage{float}
\newcommand{\ie}{\textit{i.e.}\ }

\newcommand{\blfootnote}[1]{%
  \begingroup
  \renewcommand\thefootnote{}\footnote{#1}%
  \addtocounter{footnote}{-1}%
  \endgroup
}

\title{Efficient Test-Time Adaptation For Robot Policies}

\author{
  Motasem Alfarra,\enskip
   Pietro Mazzaglia, \enskip
   Markus Peschl, \enskip
   Daniel Dijkman, \enskip
   Christos Louizos \\
  Qualcomm AI Research, Amsterdam, the Netherlands
  \\
  \texttt{malfarra@qti.qualcomm.com}
}\vspace{-0.2cm}
\begin{document}

\maketitle

\begin{abstract}
    \blfootnote{Qualcomm AI Research is an initiative of Qualcomm Technologies, Inc}
    Reinforcement-learning policies for robot locomotion can fail at deployment when operating conditions shift (e.g., rough or slippery terrain, payload changes, sensing degradation), and retraining or manual recalibration is often infeasible under on-device latency and memory constraints. 
    We study \emph{test-time adaptation} (TTA) for robotic control: adapting a pretrained policy online and self-supervised during execution, without access to rewards. Our starting point is a lightweight baseline that updates only the actor using a frozen pretrained critic as a deployment-time performance surrogate. 
    Building on this, we introduce \textsc{TEMPO}, a test-time entropy-regularized, memory-guided policy optimization method that stabilizes critic-driven updates under severe shift. \textsc{TEMPO} combines (i) an entropy-promoting regularizer on the critic’s predictive distribution to mitigate overconfident extrapolation, and (ii) a score-based memory, inspired by prioritized replay, that prioritizes low-value (“hard”) observations and applies importance sampling to focus limited computation on the most informative deployment conditions. 
    Across Go1, Go2, and G1 in MuJoCo Playground under four representative shifts and multiple architectures, \textsc{TEMPO} yields consistent gains, reaching up to 80\% improvement in a single-device setting. We further validate on a physical Unitree Go2 with an added calf payload, where fully on-device adaptation improves performance over the non-adapted baseline across real-world episodes.
\end{abstract}
\section{Introduction}

Robots deployed in the real world must contend with operating conditions that can diverge sharply from those encountered during training~\citep{kumar2021rma}.
These discrepancies arise from multiple sources: terrain variability (e.g., rocks, low-friction surfaces)~\citep{sun2025proprioceptive}, hardware changes (e.g., localized payloads or partial actuator degradation)~\citep{pouroullis2025automating}, emergent situations (e.g., a weight stuck on one arm/calf)~\citep{zhang2025disturbance,chen2023adapt}, and the broader sim-to-real gap between simulated and physical environments~\citep{peng2018sim}.
In practice, such factors rarely occur in isolation: a deployed robot may simultaneously experience altered contact dynamics, sensor noise, actuator wear, and previously unseen environmental configurations.
These distribution shifts can substantially degrade policies that perform well under nominal conditions (refer to Figure~\ref{fig:pipeline} left), making online adaptation central to robust autonomy.

A common response to combat this problem is to apply large-scale domain randomization during reinforcement learning to improve zero-shot generalization~\citep{tobin2017domain,peng2018sim}.
While effective in some cases, this approach is fundamentally constrained by what can be specified in advance: it presumes one can enumerate relevant sources of variability, model them with sufficient fidelity in simulation, and collect enough data to cover their interactions.
When robots encounter truly novel conditions, adaptation is performed offline. This interrupts operation to gather new data and fine-tune from scratch, which is misaligned with real-time, on-device compute and reliability constraints.
Architectural add-ons for adaptation can help but often increase system complexity and reduce portability across tasks, embodiments, and deployment settings~\citep{kumar2021rma}.

We study \textbf{Test-Time Adaptation} (TTA)~\citep{tent, ttt} as a practical alternative for robotic control under distribution shift.
TTA is well-suited for tackling this problem as it adapts the pretrained model at test time in an online and self-supervised manner, alleviating the need to disrupt the inference stream for retraining and requiring no labeled data.
Thus, our starting point is a simple, general baseline: we update only the actor (policy) parameters at deployment while freezing the critic, using a compact on-device memory of recent observations.
At periodic intervals during execution, the actor is adapted by maximizing the critic’s estimated action value on stored observations—treating the critic as a learned performance surrogate and requiring no online rewards.
Unlike standard deployments where the critic is used primarily for training and often discarded at inference, here we \emph{actively leverage} the pretrained critic during runtime.
While distribution shift also affects the critic, we find it still provides a sufficiently informative signal to guide test-time actor updates~\citep{chen2023adapt}; nonetheless, naive value maximization can overfit to spurious, overconfident estimates~\citep{kumar2020conservative} and under-explore the loss landscape.

In this work, and to address these limitations, we introduce an enhanced framework with two complementary elements.
First, we add an \textbf{entropy-promoting regularizer} on the critic’s predictive distribution during adaptation, which tempers overconfidence under shift and biases updates toward regions that are simultaneously high-value and high-uncertainty. This
yields more informative gradients due to the stability and exploration benefits of maximum-entropy objectives in RL~\citep{eysenbach2021maximum}, which we adapt to the \emph{critic}’s uncertainty at test time.
Second, and in the spirit of prioritized experience replay~\citep{schaul2016prioritizedexperiencereplay}, we employ a \textbf{score-based memory} that prioritizes observations inversely proportional to the critic’s predicted value, coupled with importance sampling to focus updates on the most informative deployment conditions under tight memory and latency budgets.
We evaluate on Go1, Go2, and G1 in Mujoco Playground simulation~\citep{zakka2025mujoco}, spanning quadruped and humanoid morphologies under four representative shifts (mass on torso, mass on calf, rocky terrain, slippery terrain) and three different architectures, and observe up to 50\% performance gains very efficiently in a single-device setting.
We further validate our approach on a physical Unitree \textsc{Go2} robot by attaching a $\{3, 2\}\,\mathrm{kg}$ calf payload and performing fully on-device TTA
, observing trends consistent with simulation. 
We can summarize our contributions in four folds:


\begin{figure}
    \centering
\includegraphics[width=0.33\linewidth]{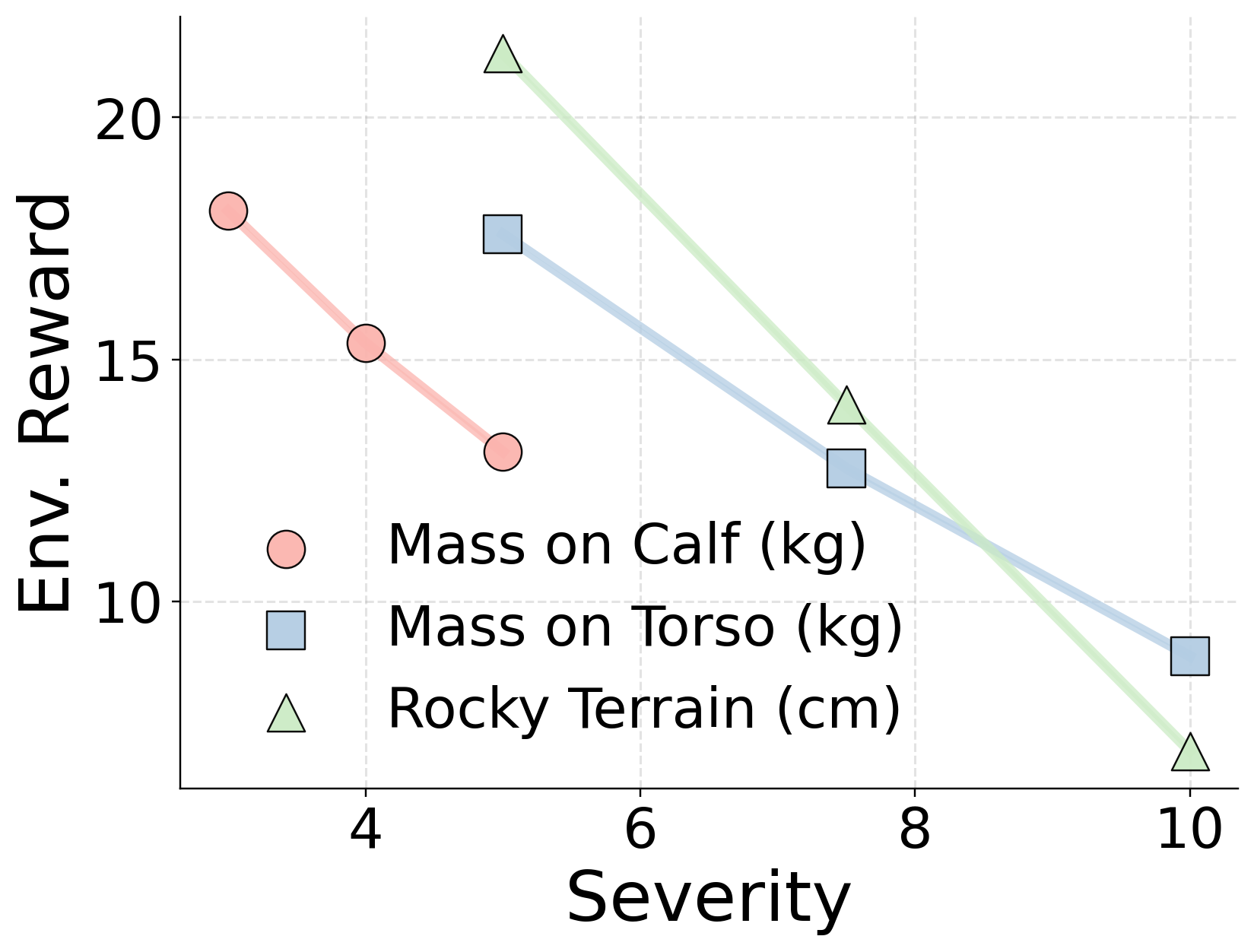}
\includegraphics[width=0.33\linewidth]{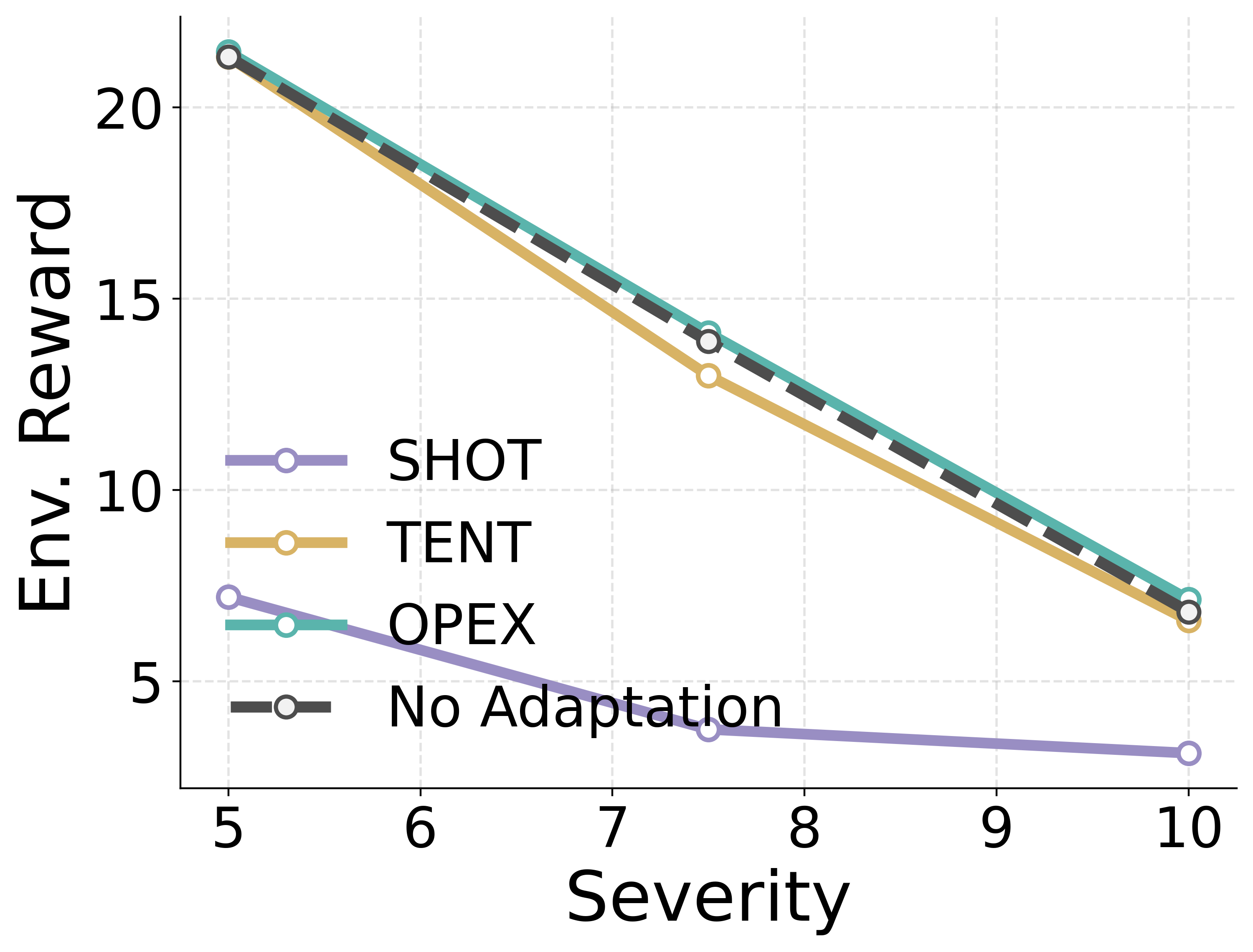}
\includegraphics[width=0.3\linewidth, trim=25 0 25 0]{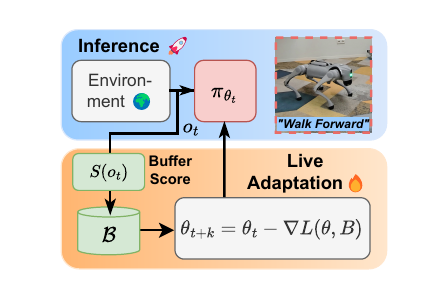}
\vspace{-2mm}
\caption{\textbf{Motivation and pipeline.} Left: Pretrained locomotion policies degrade as shift severity increases. Middle: Off-the-shelf TTA and prior RL adaptation baselines fail under a rocky-terrain shift. Right: \textsc{TEMPO} loop: execute $\pi_{\theta_t}$, update memory $\mathcal{B}$, and periodically adapt via Eq.~\eqref{eq:tempo_ent}.}
\label{fig:pipeline}
\vspace{-0.7cm}
\end{figure}

\begin{itemize}[leftmargin=*, nosep]
  \item \textbf{A lightweight, on-device TTA framework for RL control.} We adapt only the actor parameters while freezing the critic, maintain a compact memory of recent observations, and perform periodic, self-supervised updates that fit strict control-loop latency and compute budgets.
  \item \textbf{Entropy-regularized critic-driven adaptation.} We introduce an objective that maximizes the critic’s estimated action value while promoting higher-entropy critic outputs during test-time updates—mitigating overconfident value extrapolation and preserving local action coverage; the procedure is architecture-agnostic and supervision-free.

  \item \textbf{Score-based memory with importance sampling.} We propose a memory mechanism that assigns each observation a score inversely proportional to the critic’s predicted value, stores harder samples, and employs importance sampling to focus on more informative samples under memory constraints.

    \item \textbf{Comprehensive validation in simulation and on-device.} We show consistent gains on \textsc{Go1}/\textsc{Go2}/\textsc{G1} embodiments and on four deployment shifts, and confirm the same trends on a physical Unitree \textsc{Go2} via fully on-device adaptation with a $\{3,2\}\,\mathrm{kg}$ calf payload.
\end{itemize}
\section{Robot Adaptation as Test Time Adaptation}\label{sec:ra_tta}

\subsection{Problem Formulation and Evaluation Protocol}\label{sec:problem_def}
\paragraph{RL setting and distribution shift.}
We consider locomotion as a Markov decision process (MDP)
$\mathcal{M} = (\mathcal{O}, \mathcal{A}, P, \gamma)$,
where $\mathcal{O}\subseteq\mathbb{R}^{d}$ denotes the observation space,
$\mathcal{A}\subseteq\mathbb{R}^{p}$ the action space,
$P$ the (unknown) transition dynamics over observations,\footnote{We use an observation-centric formulation since neither the robot nor our method has access to privileged states.}
and $\gamma\in(0,1)$ the discount factor.
A policy $\pi_{\theta}(a\mid o)$ maps an observation $o_t\in\mathcal{O}$
(e.g., proprioception, IMU, joint states, contact signals)
to an action $a_t\in\mathcal{A}$
(e.g., target joint positions/torques).
During training, $\pi_{\theta}$ is learned in a \emph{source} training setup, inducing a training distribution over the source observations.
At deployment, the robot operates under a \emph{shifted} test-time distribution (e.g., altered friction, payload changes, rocky terrain),
which can be viewed as changes in the observations (i.e from $o$ to $\hat o$) encountered by the agent and in the induced observation-transition process (i.e. from $P$ to $\hat P$).
We assume the deployed policy is pretrained using an actor--critic framework \citep{haarnoja2018sac, fujimoto2018td3}.
In particular, alongside the pretrained policy $\pi_{\theta_0}$, training produces a critic $Q_{\psi_0}(o,a)$ that estimates the action-value of executing action $a$ when observing $o$.
In contrast with the common asymmetric setup \citep{nahrendra2023dreamwaq} where the critic is fed privileged information from the underlying simulator state, we consider a critic that consumes the same set of observations as the actor.

\textbf{Test-time adaptation as an online stream problem.}
At test time, the robot receives a stream of unlabeled observations $\{o_t\}_{t\ge 0}$ governed by the actions taken by the policy $\pi_{\theta_t}$ and the transition dynamics $\hat P$.
A Test-Time Adaptation (TTA) method is an update rule $g$ that uses only quantities available online at inference time 
to update either the model parameters, the received observation, or both.
This formulation is intentionally general: $g$ may adapt the policy parameters $\theta$, the critic parameters $\psi$, both, or a subset of either.

\textbf{Online evaluation protocol.}
Following the online/streaming evaluation paradigm~\citep{cai2021online,ghunaim2023real,alfarra2024evaluation}, we simulate the interaction between the deployed agent and the test stream at each time step $t$ as:
\begin{tcolorbox}[boxsep=1pt,left=2pt,right=2pt,top=1.5pt,bottom=1pt]
\begin{enumerate}
    \item The environment reveals an observation $o_t\sim \hat P(o_{t-1}, a_{t-1})$.
    \item The agent selects and executes an action $a_t \sim \pi_{\theta_t}(o_t)$.
    \item The adaptation rule updates parameters:
    $(\theta_{t+1}, \psi_{t+1}) \leftarrow g(\theta_t, \psi_t)$.
\end{enumerate}
\end{tcolorbox}

Notably, this protocol does not assume access to rewards or labels at test time. Although rewards are not used by the adaptation mechanism, performance is evaluated using task-level metrics computed offline or by the benchmark (e.g., episodic return in simulation, distance traveled, tracking error, stability/fall rate, task success).
These metrics are used solely for reporting and comparison, not as supervision for adaptation.

\subsection{Constructing a Baseline}\label{sec:baseline}
Test-time adaptation for robot control is challenging for two reasons. 
\textbf{(i)} As opposed to massively parallel simulation-based training \citep{li2023parallelq}, deployment typically occurs in a single environment with limited diversity over short time horizons. Consecutive observations are strongly correlated and provide a narrow view of the shifted dynamics.
\textbf{(ii)} The distribution shift can be abrupt (e.g., payload changes or slippery terrain), yet the adaptation budget is small and must fit within on-device latency constraints.
These constraints make standard TTA baselines (e.g., Tent~\citep{tent}) as well as RL-based alternatives (e.g., OPEX~\citep{park2024value}) ineffective in our setting (see Fig.~\ref{fig:pipeline}, middle)
For simplicity, we assume throughout that the policy is deterministic (\ie $a_i = \pi_{\theta_t}(o_i)$).

\textbf{Critic-guided policy update \textsc{QMax}}
One naive update strategy is to utilize the critic's action value as a surrogate self-supervised performance measure.
With that, one could potentially update the policy parameters (\ie $\theta$) to maximize the performance measure on received observations at test-time (\ie minimizing $\mathcal L = -Q_{\psi_0}\!\big(o_i, a_i\big)$).
While intuitive, the aforementioned factors (\textbf{i, ii)} make this naive per-step updates brittle and motivate aggregating evidence over time via a small memory buffer.

\textbf{FIFO buffer.}
We maintain a bounded on-device FIFO buffer $\mathcal{B}$ of capacity $N$ that stores the most recent observations.
At each time step, we append $o_t$ and evict the oldest element if $|\mathcal{B}|>N$.
Every $K$ steps, we sample a minibatch $\{o_i\}_{i=1}^{B}$ uniformly from $\mathcal{B}$ to form a more stable and diverse adaptation batch than a single observation.
We note that employing an episodic memory for TTA was also proposed by~\cite{yuan2023robust}.

\textbf{Critic-guided policy update with buffer (\textsc{QMax}).}
Given a pretrained actor--critic pair $(\pi_{\theta_0}, Q_{\psi_0})$, we define a simple self-supervised adaptation objective is to  maximize the critic’s predicted action value on buffered observations:
\begin{equation}
\mathcal{L}_{\textsc{QMax}}(\theta;\mathcal{B})
=
-\frac{1}{B}\sum_{i=1}^{B} Q_{\psi_0}\!\big(o_i, a_i\big),
\qquad a_i = \pi_{\theta}(o_i).
\label{eq:qmax_fifo}
\end{equation}
We then perform a gradient step $\theta_{t+1} \leftarrow \theta_t - \eta \nabla_{\theta}\mathcal{L}_{\textsc{QMax}}|_{\theta=\theta_t}$ with step size $\eta>0$ while freezing the critic's parameters.
While distribution shift can affect the critic’s calibration, we empirically find that $Q_{\psi_0}$ still provides a useful surrogate signal for guiding test-time updates in practice.
This baseline is lightweight and fully self-supervised, requiring only test-time observations and critic predictions (no rewards or labels).
Nevertheless, we found that \textsc{QMax} is pretty effective in recovering significant portion of the performance drop at test time.

\section{Test-time Entropy regularized, Memory guided Policy Optimization}\label{sec:our_method}
In the previous section, we formulated online test-time adaptation for a pretrained policy under distribution shift and introduced \textsc{QMax}, which adapts the policy by maximizing a pretrained critic on a small buffer of recent observations.
As we demonstrate empirically in Section~\ref{sec:main_results}, \textsc{QMax} can be effective, but it may become unstable under severe shifts: when the critic is queried out of distribution, directly ascending $Q_{\psi_0}(o,a_t)$ may yield overly sharp updates, reduced action coverage, and brittle closed-loop behavior.
We propose \textbf{TEMPO} (\textit{Test-time Entropy-regularized, Memory-guided Policy Optimization}), which improves stability and sample-efficiency by combining two lightweight modifications: 
(i) An entropy-regularized critic objective, and (ii) a score-based memory that prioritizes informative test-time observations.
As described in Section~\ref{sec:problem_def}, TEMPO updates only the policy parameters $\theta$ at test time.


\subsection{Entropy Maximization}\label{sec:mcent}
While action perturbation broadens the local neighborhood explored by each update, it does not explicitly discourage overconfident critic-driven collapse.
TEMPO therefore augments critic maximization with an entropy term defined on the critic's predictive distribution, encouraging updates that are high-value yet avoid overly peaky solutions.

\paragraph{Distributional critic and entropy.}
Following~\cite{seo2025fasttd3}, we assume the actor--critic pretraining provides a distributional critic
$Z_{\psi_0}(o,a)\in\mathcal{P}(\mathcal{Z})$ where $\mathcal{P}(\mathcal{Z})$ is the probability simplex over a discretized value support $\mathcal{Z}$, with
$Q_{\psi_0}(o,a) = \mathbb{E}_{z\sim Z_{\psi_0}(o,a)}[z]$.
To avoid the high confidence regions over the critic's prediction we propose an \textbf{Entropy-regularized objective.}
TEMPO optimizes the following entropy-regularized loss on buffered observations with:
\begin{equation}
\mathcal{L}_{\textsc{TEMPO}}(\theta;\mathcal{B})
=
-\frac{1}{B}\sum_{i=1}^{B}
\left[
Q_{\psi_0}\!\big(o_i,  a_i\big)
\;+\;
\lambda\, \mathcal H\!\left(Z_{\psi_0}(o_i, a_i)\right)
\right],
\label{eq:tempo_ent}
\end{equation}
where $\lambda>0$ controls the strength of entropy regularization.
The entropy term tempers overconfident maximization by favoring updates that remain robust in regions where the critic's predictions are less concentrated, improving stability under large shifts.

\subsection{Score-Based Memory}\label{sec:score_memory}
Next, we ask the following question: given the memory and compute constraints for on-device adaptation, which samples shall we store? And which one shall we adapt on?
FIFO buffers treat all recent observations as equally useful, even though under distribution shift only a subset of encountered states may be responsible for performance degradation (e.g., slip events or destabilizing contacts).
TEMPO therefore replaces FIFO with a score-based memory that preferentially stores and samples the most informative observations. 

\textbf{Scoring and buffer maintenance.}
Given the current policy and pretrained critic, we assign each observation a score
\begin{equation}
S(o) = \frac{1}{Q_{\psi_0}\!\big(o,\pi_\theta(o)\big)+\varepsilon},
\label{eq:score}
\end{equation}
with a small constant $\varepsilon>0$ for numerical stability.
High scores correspond to low predicted value and thus ``hard'' states.
TEMPO maintains a buffer $\mathcal{B}$ of capacity $N$ storing pairs $(o, S(o))$.
When a new observation $o_t$ arrives, we compute $S(o_t)$ and insert it if it exceeds the minimum score currently in the buffer; in that case, the lowest-scoring element is evicted.
This yields a compact memory that tracks the most challenging deployment observations under the current policy.

\textbf{Importance sampling for adaptation.}
During each adaptation step, TEMPO samples a minibatch from $\mathcal{B}$ using probabilities proportional to the stored scores:
\begin{equation}
p_j \;=\; \frac{\hat{S}_j}{\sum_k \hat{S}_k}, \qquad\qquad \text{where} \,\, \hat{S}_j = S_j - \min_{i}S_i + \varepsilon
\label{eq:importance_sampling}
\end{equation}
where $p_j$ is the probability of sampling the $j^{th}$ element from the memory, and $S_j$ is computed according to Eq.~\eqref{eq:score}. TEMPO then draws $B$ observations from $\mathcal B$ without replacement according to $\{p_j\}$.
This focuses limited on-device adaptation compute on states that are most associated with degraded performance, improving sample-efficiency compared to uniform FIFO sampling.

\textbf{Putting All Together.}
TEMPO combines three lightweight ingredients into a single on-device adaptation loop.
At each control step, the robot executes the current deterministic policy $a_t=\pi_{\theta_t}(o_t)$ and records the observation in a bounded memory.
Unlike FIFO, TEMPO maintains a score-based buffer that preferentially keeps ``hard'' observations (Eq.~\eqref{eq:score}) and samples adaptation minibatches via importance sampling (Eq.~\eqref{eq:importance_sampling}).
Every $K$ steps, TEMPO performs a single gradient update on $\theta$ (with $Q_{\psi_0}$ frozen) by minimizing the entropy-regularized objective in Eq.~\eqref{eq:tempo_ent}.
Figure~\ref{fig:pipeline} (right) summarizes the pipeline.
A detailed algorithm for adapting with TEMPO is in Appendix~\ref{sec:app_algos}.

\section{Related Work}
\paragraph{Domain Shifts on Robots}
Robots deployed outside controlled training conditions face \emph{domain shifts} such as changes in contact dynamics, inertial properties, and sensing, which can substantially degrade locomotion policies. A common response is to harden policies in simulation via accurate actuator modeling \citep{Tan-RSS-18,hwangbo2019learning,zhu2025learning,shin2025reinforcement}, domain randomization \citep{peng2018sim,joonho2020learning} and massively parallel training \citep{rudin2022learning,zakka2025mujoco}. This improves zero-shot robustness within the training envelope. To better handle unstructured environments, many approaches additionally incorporate exteroception (e.g., elevation maps or images) to anticipate terrain-induced shifts \citep{takahiro2022learning,lai2023sim,he2025attention,yu2024learning,dong2025marg}, or auxiliary mechanisms (e.g., asymmetric training, distillation, learned estimators) that infer latent factors \citep{chen2020learning,kumar2021rma,ji2022concurrent,suyoung2023deformable}. However, when deployment conditions drift beyond the training distribution, these approaches often still require retraining or additional structure, motivating lightweight \emph{deployment-time} adaptation under strict hardware constraints.
Adapting at deployment time has been explored in adjacent settings: ~\cite{park2024value} analyze offline RL bottlenecks and propose simple test-time policy improvement methods, while ~\cite{chen2023adapt} adapt on-the-fly in single-life deployment by selecting and modulating a \emph{repertoire} of pretrained behaviors using value estimates. 
In contrast, we target lightweight \emph{on-device} test-time adaptation for locomotion under strict latency and memory budgets, by adapting a \emph{single} deployed actor via critic-guided updates.

\paragraph{Test-Time Adaptation}
Test-Time Adaptation (TTA) mitigates the performance drop of pretrained models under test-time distribution shift by adapting at inference using only unlabeled test data~\citep{mancini2018kitting,cfa}. Methods either update model parameters online~\citep{shot,tent} or adapt the test input to better match the training distribution~\citep{dda}. Early approaches focused on updating Batch Normalization statistics~\citep{adabn}, while later work introduced stronger self-supervised objectives, including entropy minimization and stabilized variants~\citep{tent,eata,sar}, information maximization~\citep{shot}, and consistency/teacher--student updates~\citep{yuan2023robust}, as well as auxiliary-task-driven adaptation~\citep{alfarra2025testtime}. Most progress has been demonstrated on perception benchmarks—primarily classification under synthetic corruptions and covariate shift~\citep{imagenetc}—with fewer works on structured prediction~\citep{actmad} or other modalities such as missing modality~\citep{ramazanova2024test}. Translating TTA to robotics is challenging because adaptation must preserve stable closed-loop control under strict latency/memory budgets and highly correlated on-policy streams; our work addresses this by tailoring lightweight, critic-guided TTA to on-device locomotion.

\section{Experiments}\label{sec:experiments}
\textbf{Environments and pretrained policies.}
We evaluate TTA on three simulated locomotion tasks from MuJoCo Playground~\citep{zakka2025mujoco}: \textsc{Go1-joystick} (quadruped), \textsc{Go2-joystick} (quadruped), and \textsc{G1-joystick} (humanoid).
Since Section~\ref{sec:ra_tta} places no assumptions on the policy architecture, we consider three representative pretrained policy families:
\textsc{Fast-TD3}~\citep{seo2025fasttd3}, \textsc{SimbaV2}~\citep{lee2025simbav2}, and  Rapid Motor Adaptation~(\textsc{RMA})~\citep{kumar2021rma}. 

\textbf{Training protocol and distribution shifts.}
To disentangle test-time robustness from training-time regularization, we train $\pi_\theta$ on a \emph{flat} terrain \emph{without} domain randomization, and evaluate (and adapt) under a single fixed test environment that emulates a realistic deployment condition.
We consider four shift types parameterized by a severity scalar $\delta$:
\textbf{(i)} \emph{Rough terrain}, with height variation $\delta$;
\textbf{(ii)} \emph{Calf mass perturbation}, attaching a mass of $\delta$ kg to one calf;
\textbf{(iii)} \emph{Torso mass perturbation}, attaching a mass of $\delta$ kg to the torso; and
\textbf{(iv)} \emph{Slippery terrain}, setting the ground friction coefficient to $1-\delta$.
Unless otherwise stated, each test episode contains a \emph{single} shift type (i.e., no mixed corruptions).


\textbf{Adaptation protocol.}
Across all methods, we use Adam with learning rate $5\times10^{-6}$.
We adapt all learnable parameters of $\pi_\theta$ while freezing the critic(s).
We maintain a replay buffer of size $|\mathcal{B}|=16\text{k}$ and use batches of size $B=2\text{k}$.
Note that with 49-dimensional observations and $|\mathcal{B}|=16\text{k}$, the FIFO buffer footprint is merely $\approx 3$~MiB; score-based variants require $\approx 3.05$~MiB.
Adaptation is performed online with period $K=128$: every $K$ environment steps, we sample a batch from $\mathcal{B}$, compute gradients, and take a single optimizer step.
For \textsc{TEMPO}
we set the entropy weight to $\lambda=0.5$.
In the following sections, we will provide a comprehensive evaluation demonstrating the benefits of TTA under distribution shift and the consistent performance gains of \textsc{TEMPO}.
Unless stated otherwise, we fix the policy architecture to Fast-TD3 and the environment to Go2 on Flat terrain without domain randomization. 
Further experimental details can be found in Appendix~\ref{sec:app_exps_setup}.



\begin{figure}
    \centering
    \hspace{2cm}\includegraphics[width=0.4\linewidth]{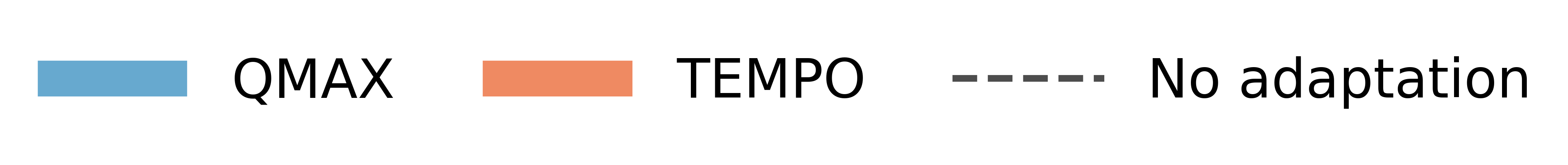}
    \newline
    \includegraphics[width=0.245\linewidth]{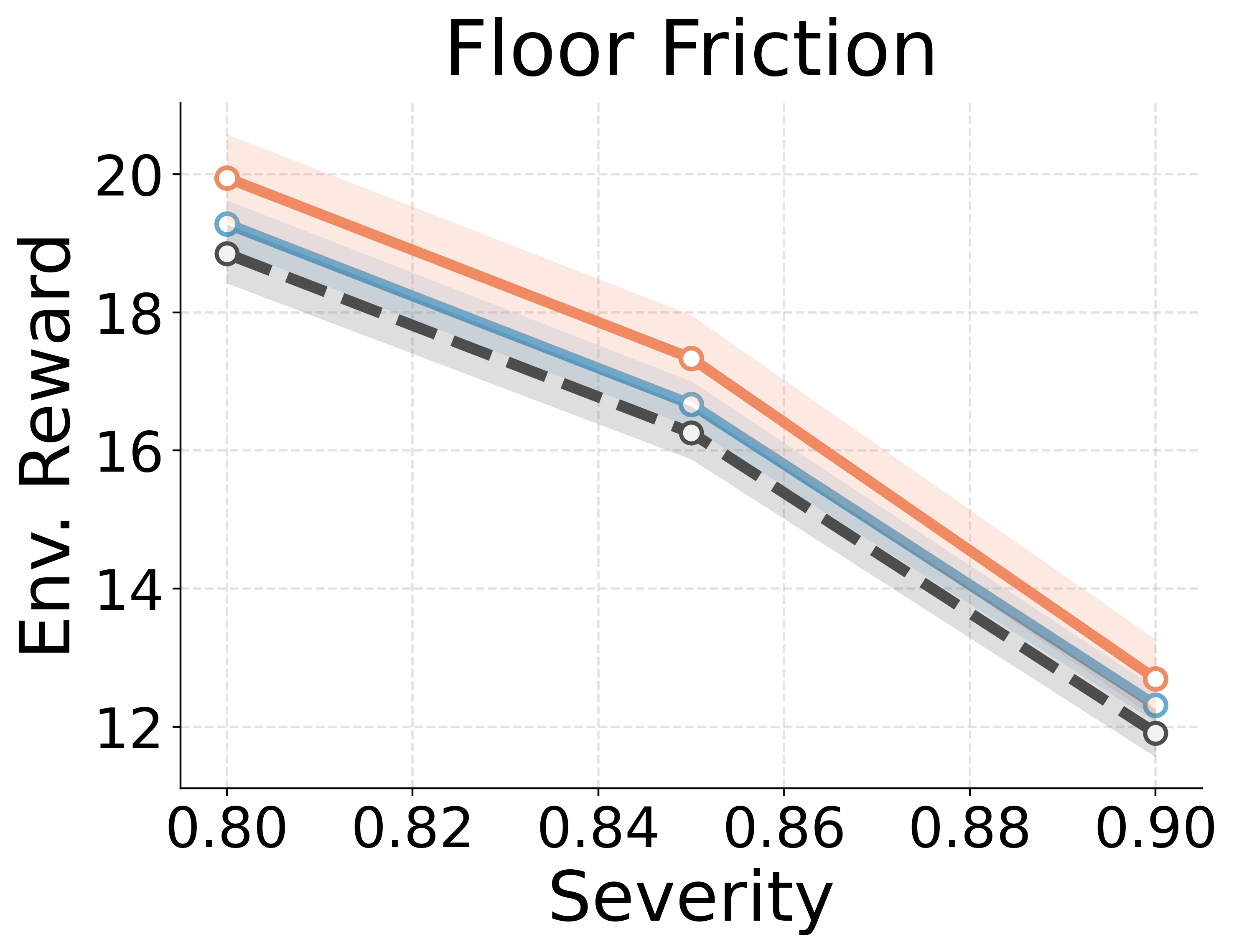}
    \includegraphics[width=0.245\linewidth]{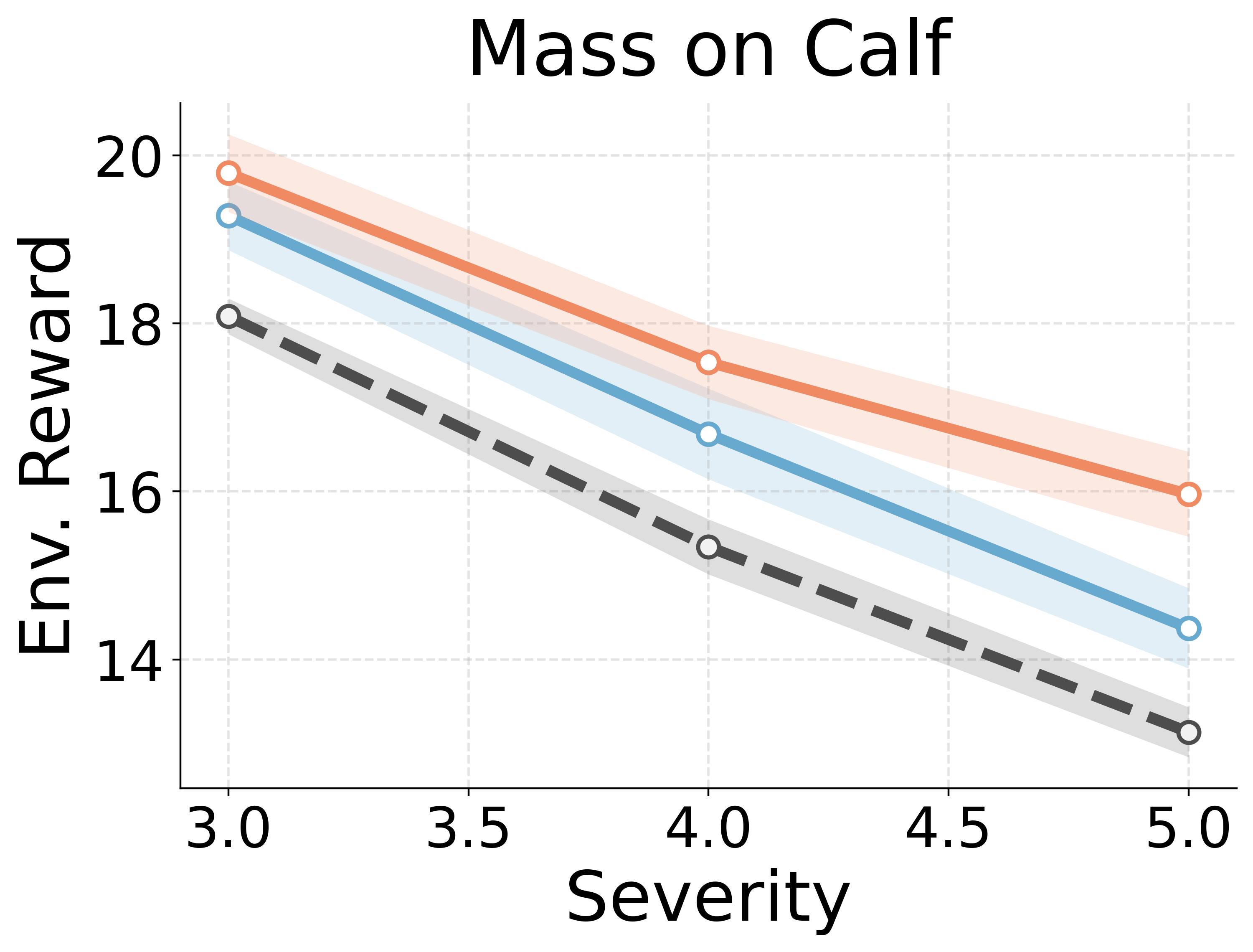}
    \includegraphics[width=0.245\linewidth]{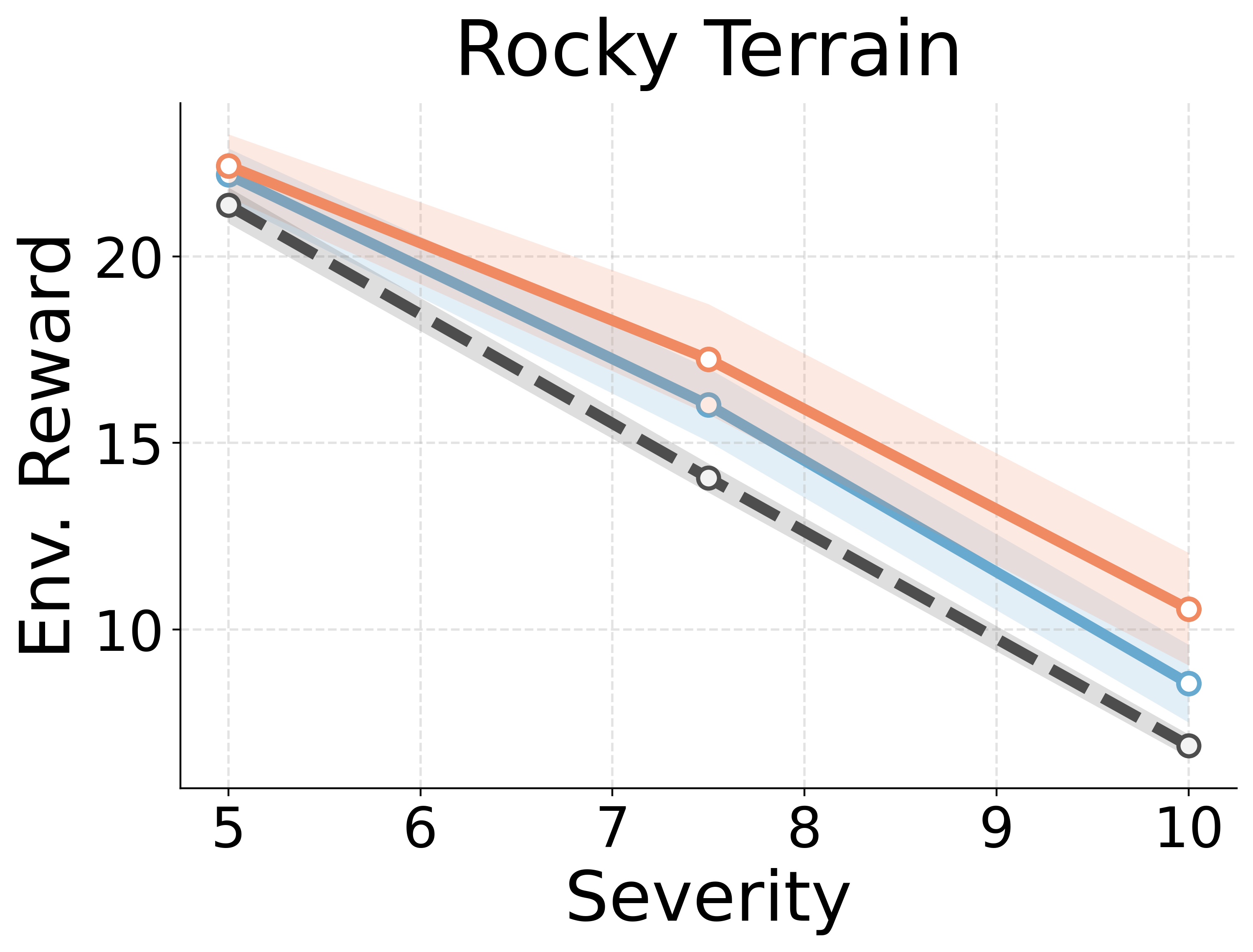}
    \includegraphics[width=0.245\linewidth]{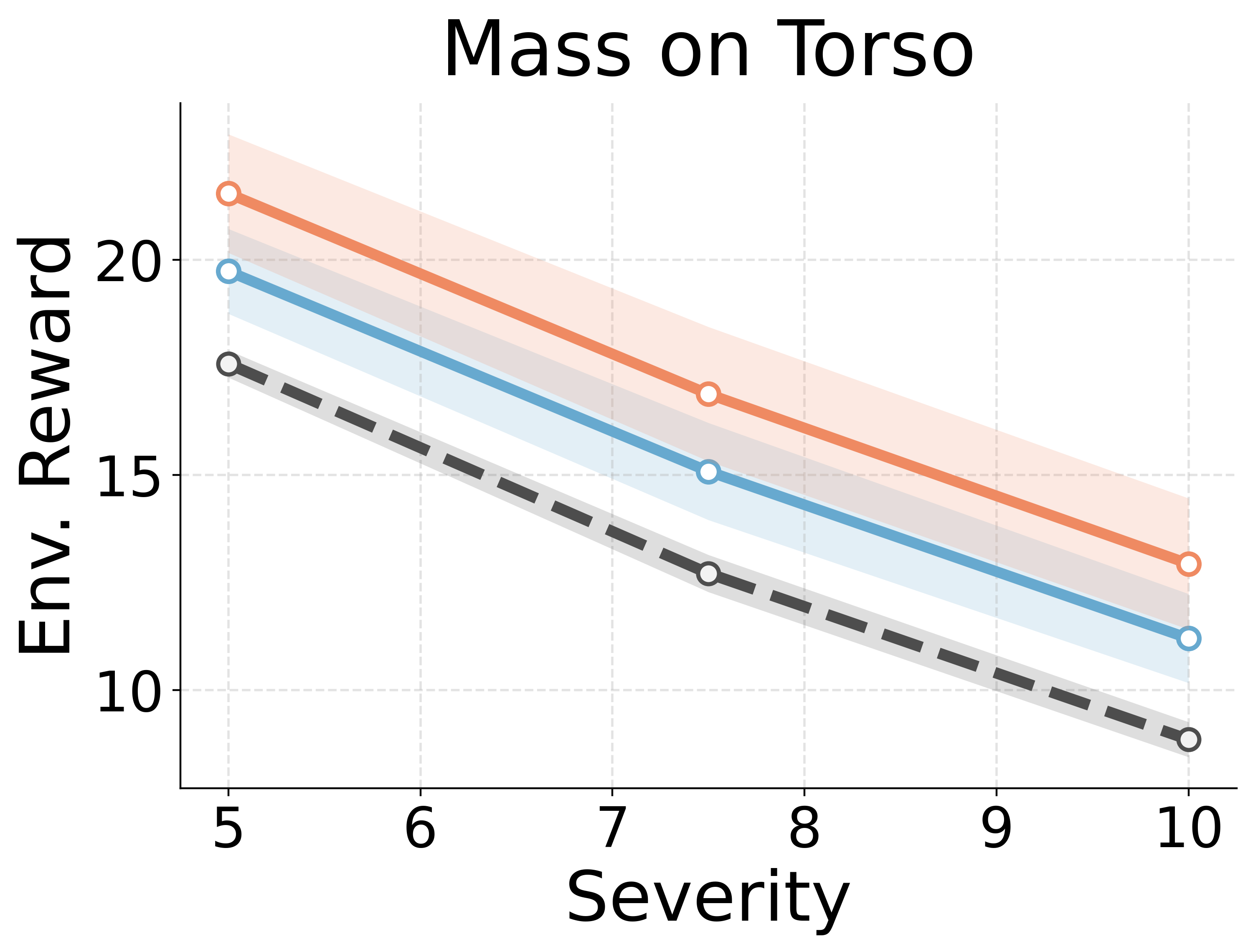}
    \includegraphics[width=0.245\linewidth]{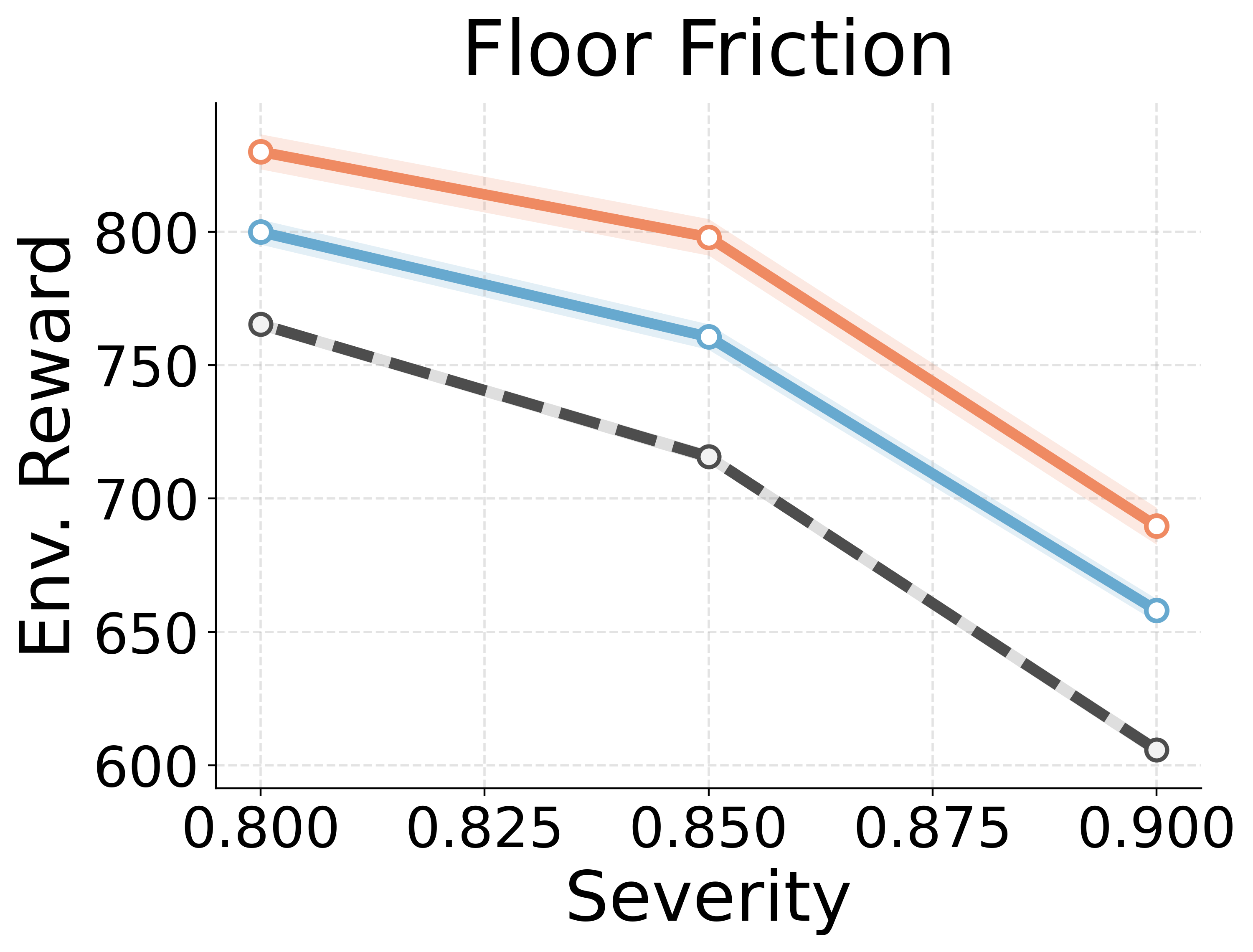}
    \includegraphics[width=0.245\linewidth]{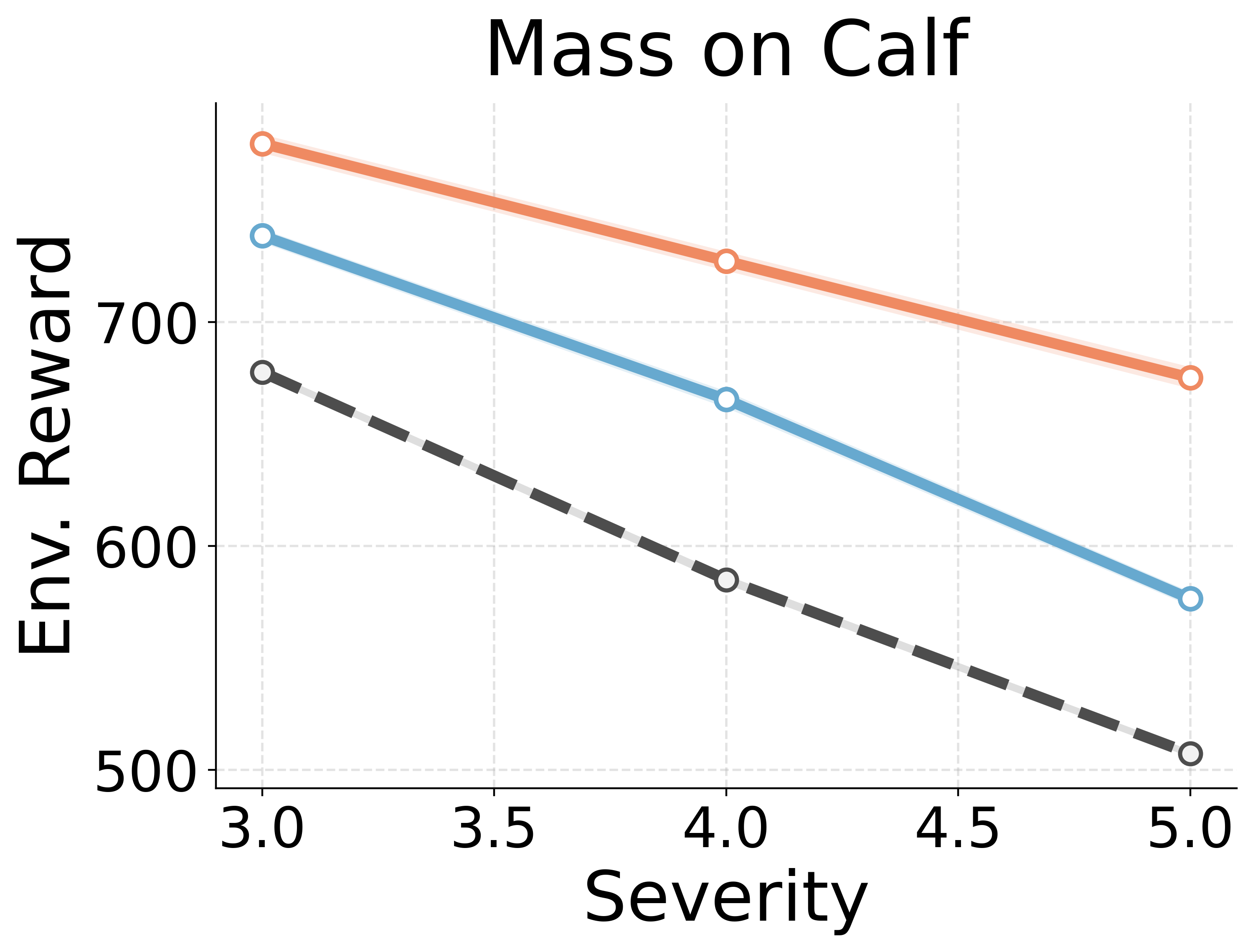}
    \includegraphics[width=0.245\linewidth]{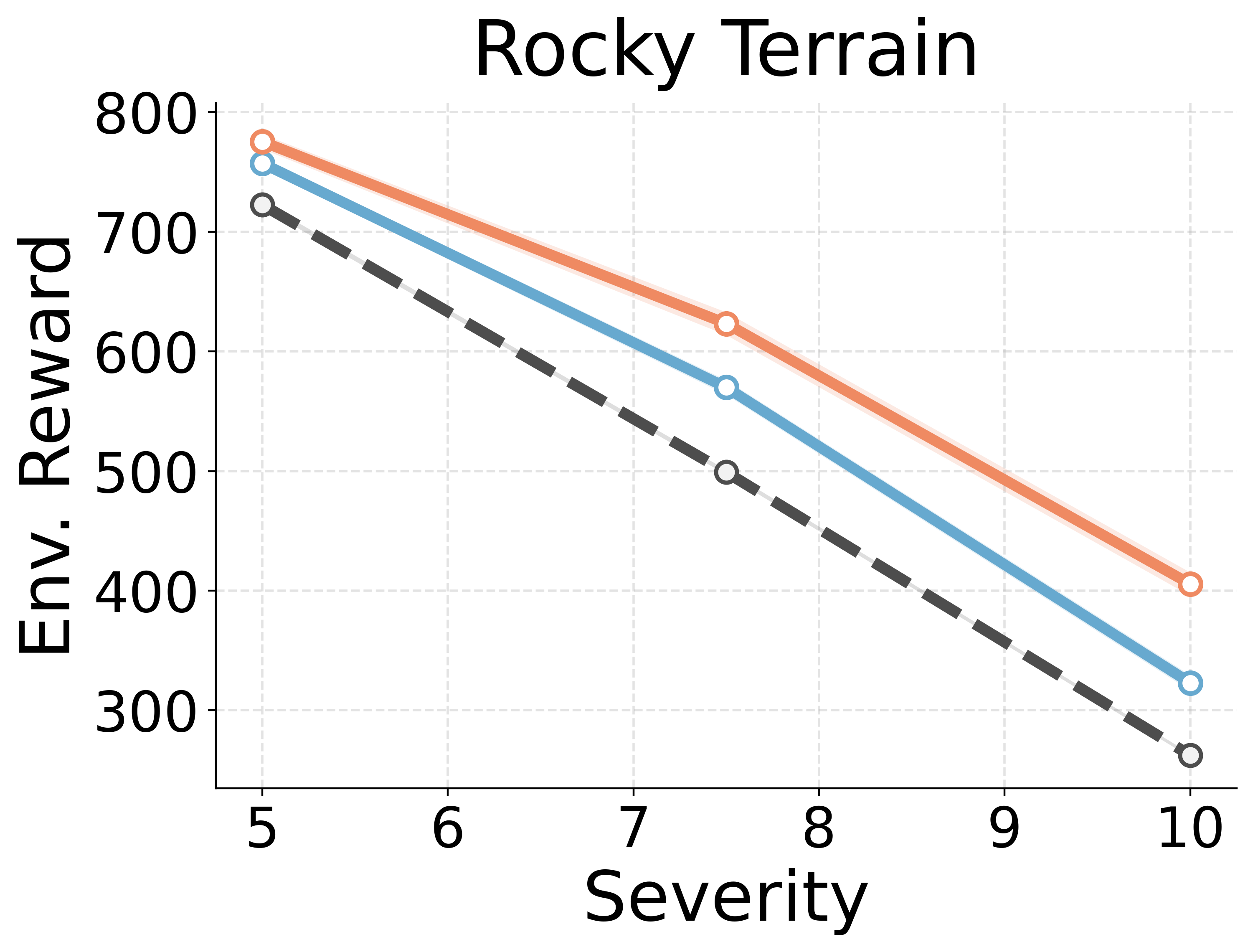}
    \includegraphics[width=0.245\linewidth]{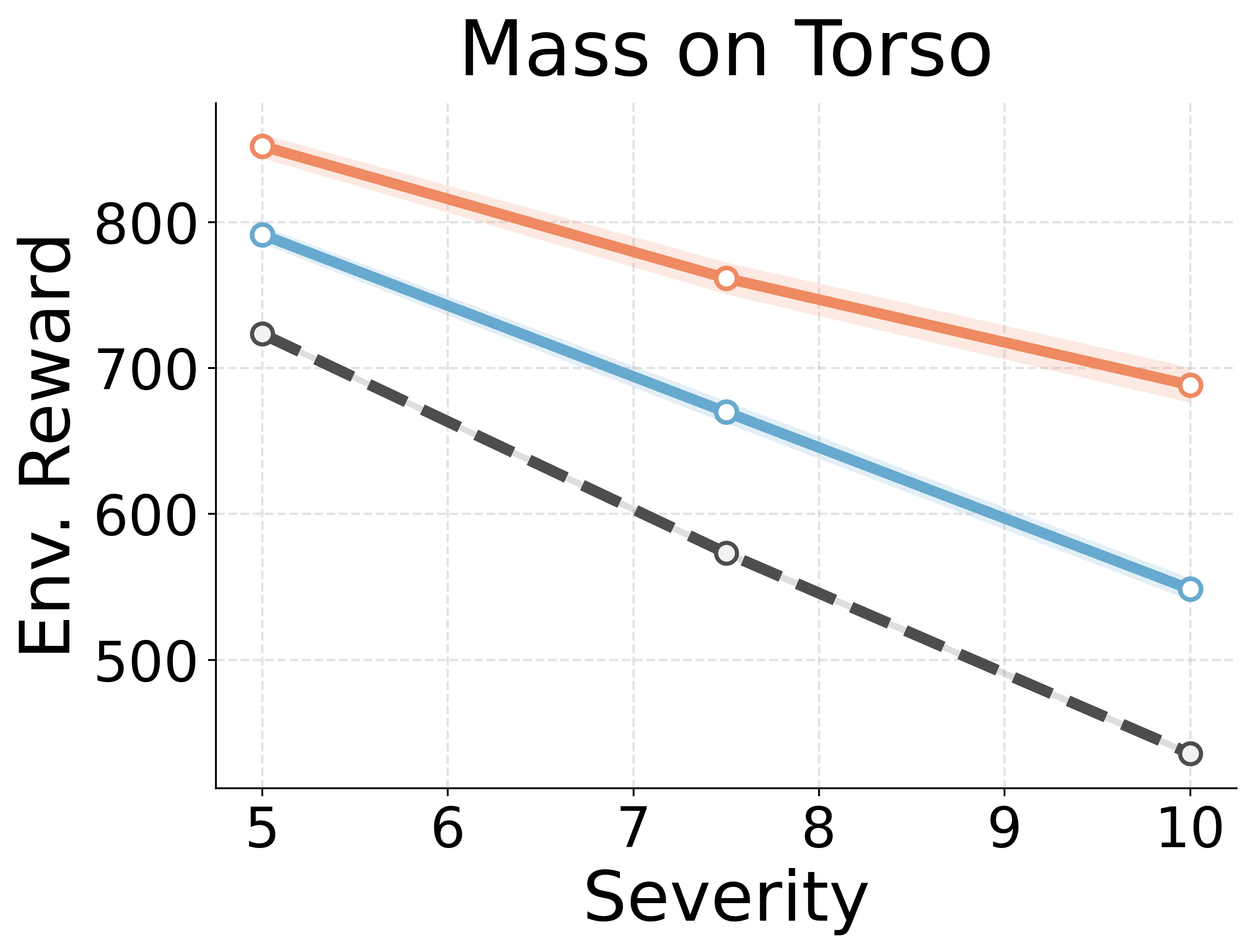}
    \caption{\textbf{Main results on Go2 under deployment-time distribution shift.}
Top: episodic return. Bottom: average forward linear velocity.
Shifts (left$\rightarrow$right): $1-$friction, calf mass, rough terrain, torso mass (severity increases along x-axis).
As the severity of the domain shift increases, performance degrades more rapidly.
Naive adaptation via QMAX recovers part of the performance drop.
Our proposed \textsc{TEMPO} consistently provides the largest performance gain, especially under severe shifts.}\vspace{-0.35cm}
\label{fig:main_results}
\end{figure}

\subsection{Effectiveness of TEMPO}\label{sec:main_results}
We first evaluate whether lightweight \emph{test-time adaptation} (TTA) can recover locomotion performance under realistic deployment shifts, and whether \textsc{TEMPO} provides robust gains beyond critic maximization and general-purpose TTA baselines. 
To do so, we test on four shift families with increasing severity: slippery terrain, calf mass perturbation, rough terrain, and torso mass perturbation. Adaptation is performed online during execution using only test-time observations;
rewards are never used for updating and are only used for offline evaluation. 
All results are averaged across six seeds.

Figure~\ref{fig:main_results} reports \textbf{episodic return} (top row) and \textbf{average forward linear velocity} (bottom row), capturing both task-level performance and the quality of locomotion. Across all shifts, the pretrained policy without adaptation degrades rapidly as severity increases, indicating that even strong policies trained under nominal conditions are brittle when contact dynamics or inertial properties change at deployment. Introducing TTA reliably mitigates this drop: the simple critic-guided baseline \textsc{QMAX} recovers performance across a wide range of severities, confirming that a frozen critic can provide an informative self-supervised signal at test time. However, \textsc{QMAX} often saturates under harder shifts.
In contrast, \textsc{TEMPO} is consistently best across all shift families and throughout the severity range, with the clearest advantages emerging in the most challenging regimes. Importantly, these gains are mirrored in forward velocity: \textsc{TEMPO} maintains faster, more continuous locomotion under shift, suggesting that improvements reflect better stabilized behavior rather than reward-only artifacts.

\begin{figure}[t]
    \centering
    \begin{minipage}{0.5\linewidth}
        \centering
        \captionof{table}{\textbf{Real-world deployment on Unitree Go2 under a calf-mass shift.}
\textsc{TEMPO} significantly improves average reward over the non-adapted baseline for various input commands.}
        \label{tab:tta_vs_base}
        \begin{tabular}{ c c c c}
            \toprule
              $v_x$ & $w_z$ & No adaptation & TEMPO \\
            \midrule
              0.5  & 0.0  & $93.2 \pm 1.3$ & $\mathbf{95.2 \pm 0.6}$ \\
             -0.5 & 0.0  & $98.7 \pm 1.6$ & $\mathbf{113.5 \pm 0.7}$ \\
             -0.7 & 0.0  & $88.1 \pm 1.1$ & $\mathbf{96.5 \pm 1.0}$\\
            0.7  & 0.6  & $88.4 \pm 3.3$ & $\mathbf{101.6 \pm 1.4}$\\
             0.7  & -0.6 & $80.5 \pm 1.0$ & $\mathbf{86.9 \pm 1.2}$ \\
        \midrule
\multicolumn{2}{c}{Average} 
& $89.8$ 
& $\mathbf{98.7}$ \\        \bottomrule
\end{tabular}
    \end{minipage}
    \hfill
    \begin{minipage}{0.45\linewidth}
        \centering
        \captionof{figure}{\textbf{Real-world distribution shift on Unitree Go2.}
Weight plates mounted on a calf to induce an inertial perturbation during on-device adaptation.}
        \includegraphics[width=0.9\linewidth,
        trim=2cm 14cm 4cm 19cm,
        clip]{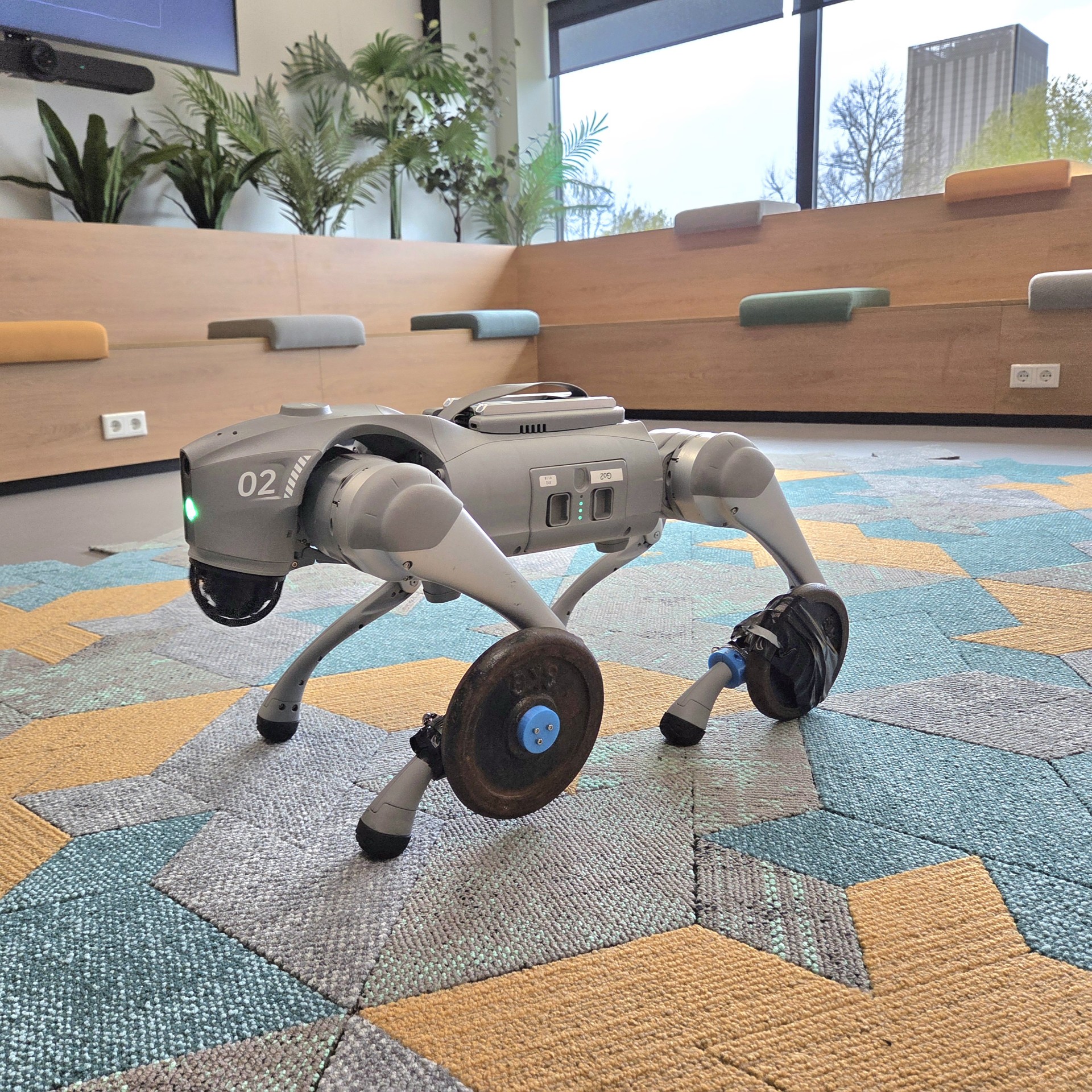}

        \label{fig:go2_weights}
    \end{minipage}
    \vspace{-0.4cm}
\end{figure}

\subsection{Real World Deployment}\label{sec:real_world_deployment}

We validate \textsc{TEMPO} on a physical Unitree \textsc{Go2} quadruped to assess whether the improvements observed in simulation translate to real hardware. To induce a controlled deployment shift, we attach weight plates to both the front left and rear left calves using a custom 3d-printed rigid mount (Fig.~\ref{fig:go2_weights}), altering the leg’s inertial properties and contact dynamics while keeping the policy and controller stack unchanged. Adaptation is performed once on-device for 4 minutes and using the same streaming protocol as in simulation: we continuously update the actor parameters online while freezing the critic, using only real observations and critic predictions.

Table~\ref{tab:tta_vs_base} reports the environment reward across for various velocity commands $cmd=(v_x, w_z)$, where $v_x$ is the desired forward linear velocity and $w_z$ is the desired angular velocity. The evaluation for each command consists of 5 trials spanning a 3 second window (150 steps each). To define a useful metric for real world performance, we evaluate the undiscounted sum of the per-timestep linear-velocity tracking rewards $r_t = \exp\!\left(-\lVert \mathbf{v}^{\mathrm{cmd}}_{xy} - \mathbf{v}^{\mathrm{loc}}_{xy} \rVert^2 \right)$. Thus, each timestep contributes a value in $(0, 1]$, resulting in a maximum possible value of $150$. \textsc{TEMPO} outperforms the non-adapted baseline for each of the tested commands, improving the mean score from $89.8$ to $98.7$. 
Gains are consistent across command types, with the largest improvements under walking backwards at medium speed ($v_x=-0.5$), where \textsc{TEMPO} increases the score by $15\%$ from $98.7$ to $113.5$. 
Overall, these results provide evidence that \textsc{TEMPO} remains effective under real-world distribution shift and can improve locomotion performance in a fully on-device self-supervised setting.

\begin{table}[t]
\centering
\caption{\textbf{Performance Comparison Under Mixed Corruptions.} We report the environment reward (six seeds) for the non-adapted policy, QMAX, and our proposed TEMPO when the policy is subjected to a composition of corruptions (e.g. mass on torso and rocky terrain simultaneously). Our proposed TEMPO consistently provides the best performance by a significant margin.}
\resizebox{\linewidth}{!}{
\begin{tabular}{lcccccccc}
\toprule
& \multicolumn{4}{c}{Mass on Calf = 2\,kg} & \multicolumn{4}{c}{Mass on Torso = 5\,kg} \\
\cmidrule(lr){2-5} \cmidrule(lr){6-9}
& \multicolumn{2}{c}{Floor friction} & \multicolumn{2}{c}{Rocky terrain}
& \multicolumn{2}{c}{Floor friction} & \multicolumn{2}{c}{Rocky terrain} \\
\cmidrule(lr){2-3} \cmidrule(lr){4-5} \cmidrule(lr){6-7} \cmidrule(lr){8-9}
Method 
& 0.90 & 0.85 & 0.03 & 0.06 
& 0.90 & 0.85 & 0.03 & 0.06 \\
\midrule
No Adapt & $9.2_{\pm 0.1}$ & 12.9$_{\pm 0.2}$ & $19.9_{\pm 0.1}$ & $14.5_{\pm 0.2}$ &  4.2$_{\pm 0.2}$& 6.2$_{\pm 0.1}$ & 14.7$_{\pm 0.3}$ & 7.8$_{\pm 0.2}$ \\
QMAX    & $9.7_{\pm 0.2}$  & 13.4$_{\pm 0.4}$ & $20.7_{\pm 0.1}$ & $15.9_{\pm 0.2}$ &  4.8$_{\pm 0.1}$& 7.0$_{\pm 0.3}$  & 17.3$_{\pm 0.5}$ &9.9$_{\pm 0.2}$  \\
TEMPO   & \textbf{10.4$_{\pm 0.2}$} & \textbf{14.4$_{\pm 0.1}$} & \textbf{21.2$_{\pm 0.3}$} & \textbf{17.5$_{\pm 0.2}$} &  \textbf{5.6$_{\pm 0.2}$}& \textbf{9.0$_{\pm 0.4}$}  & \textbf{20.1$_{\pm 0.2}$} & \textbf{14.6$_{\pm 0.2}$} \\
\bottomrule
\end{tabular}
}\label{tab:mixed_corruptions}
\end{table}

\subsection{Mixed Corruption Setting  }\label{sec:mixed_corr}
Real deployments rarely involve a single isolated shift; instead, multiple factors can co-occur (e.g., payload changes together with altered contact dynamics). We therefore evaluate \emph{mixed corruptions} by combining a \textbf{payload perturbation} with a \textbf{terrain shift} within the same episode. Specifically, we attach a fixed mass either to the \textbf{calf} ($2$\,kg) or the \textbf{torso} ($5$\,kg), and simultaneously test on either \textbf{slippery terrain} ($1-$friction) or \textbf{rough terrain} (rocky heightfield), each at two severities.

Table~\ref{tab:mixed_corruptions} summarizes the results. Across \emph{all 8} mixed conditions, \textsc{TEMPO} is best: averaged over the table, it improves the environment reward from $11.18$ (\textsc{No Adapt}) and $12.48$ (\textsc{QMAX}) to $14.09$, i.e., a gain of $+2.91$ ($+26.1\%$) over \textsc{No Adapt} and $+1.61$ ($+12.9\%$) over \textsc{QMAX}. The advantage is especially pronounced for the harder torso-payload compositions, where \textsc{TEMPO} raises the average reward from $8.23$ to $12.33$ ($+4.10$, $+49.8\%$), and from $10.00$ to $12.33$ versus \textsc{QMAX} ($+2.33$, $+23.3\%$). The largest single-case improvement occurs under torso mass with rocky terrain at the highest severity, where \textsc{TEMPO} reaches $14.6$ compared to $10.5$ for \textsc{QMAX} ($+4.1$, $+39.0\%$) and $7.8$ without adaptation ($+6.8$, $+87.2\%$), demonstrating clear benefits under compounding shifts.

\subsection{Continual TTA }\label{sec:continual_tta}

\begin{figure}[b]
    \centering
    \includegraphics[width=0.32\linewidth]{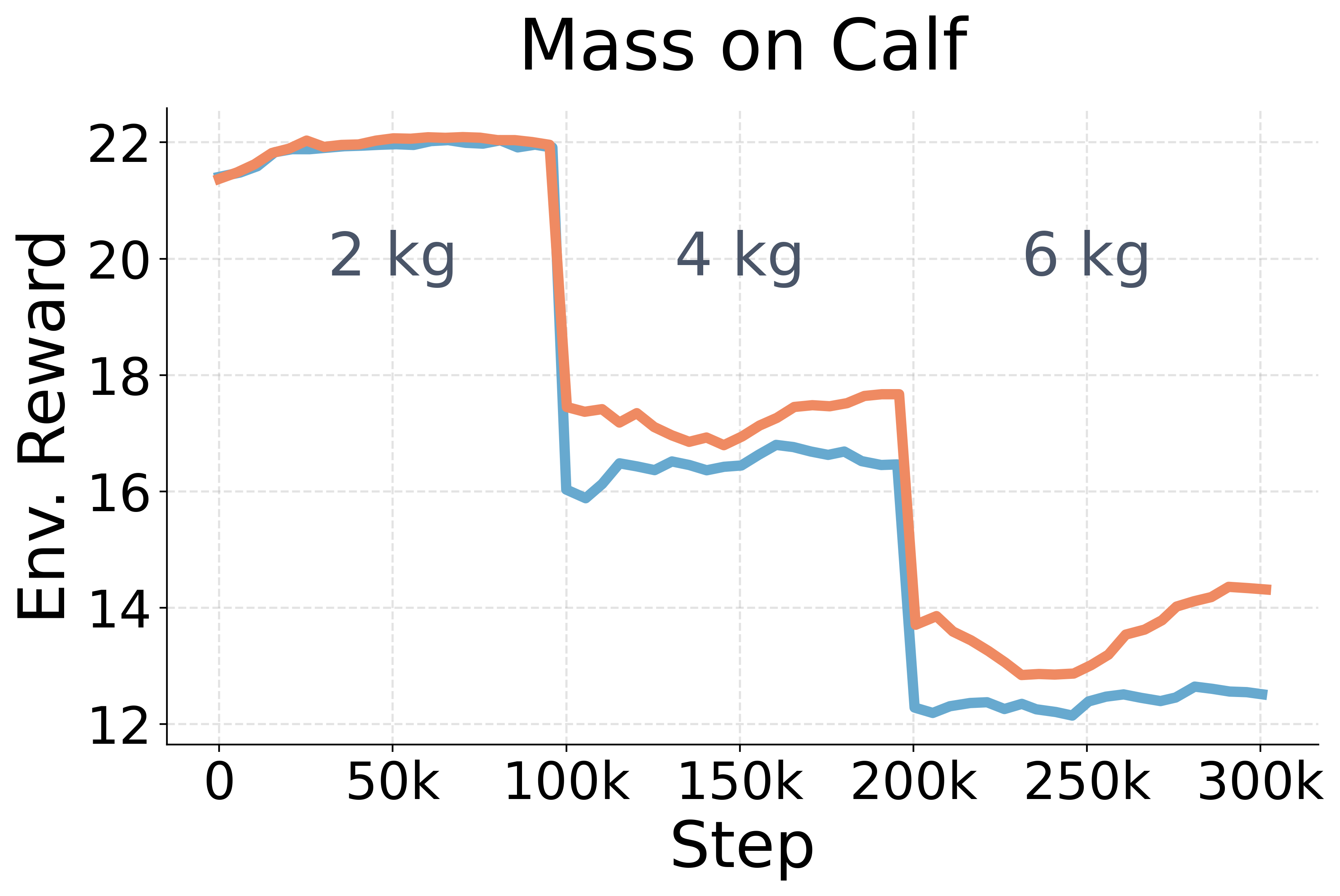}
    \includegraphics[width=0.32\linewidth]{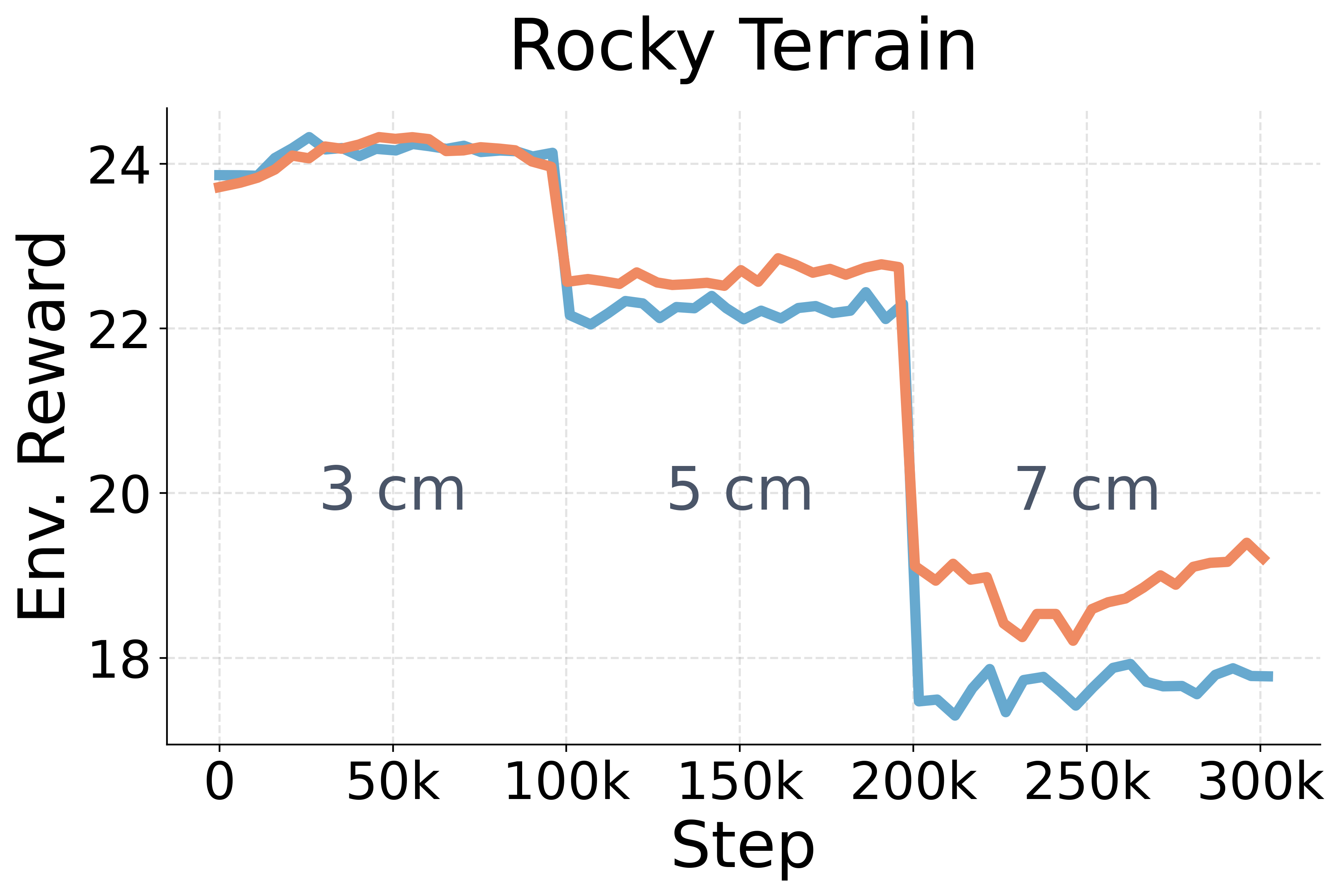}
    \includegraphics[width=0.32\linewidth]{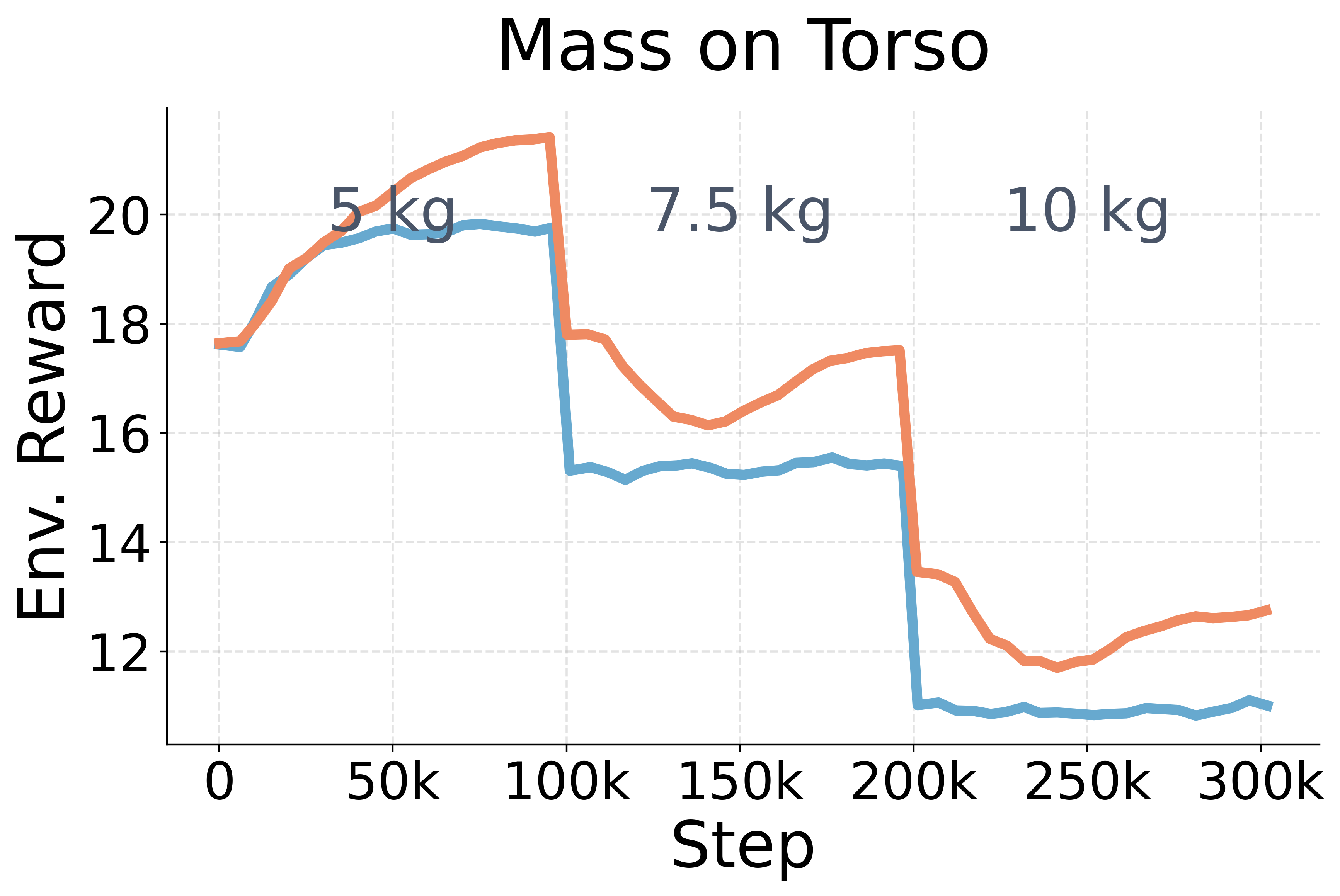}
    \caption{\textbf{Continual TTA under progressively increasing shift severity.}
We report environment rewards versus adaptation steps as the deployment shift intensifies over time.
\textsc{TEMPO} recovers faster and converges to a higher performance plateau than \textsc{QMAX} across all corruptions.}
    \label{fig:continual_tta}
\end{figure}

So far, we evaluated adaptation under a fixed deployment condition. In practice, however, distribution shift is often \emph{non-stationary}: terrain properties can change gradually, payloads can vary over time, and contact dynamics can drift during prolonged operation. We therefore consider a \emph{continual TTA}~\citep{cotta} setting in which the deployed policy experiences a sequence of increasingly difficult conditions, implemented by continually increasing the severity of the domain shift during evaluation.

Figure~\ref{fig:continual_tta} reports average return as adaptation progresses in this non-stationary stream. The results highlight a clear separation between methods: \textsc{TEMPO} exhibits faster recovery after the shift intensifies and reaches a higher performance plateau than \textsc{QMAX}. This gap is most evident as the stream becomes more challenging, where naive critic maximization improves initially but saturates earlier, while \textsc{TEMPO} continues to make progress and maintains stronger steady-state performance. Overall, these results indicate that \textsc{TEMPO} is not only effective under a fixed test condition, but also well-suited to \emph{continual} deployment where the test distribution evolves over time and adaptation must remain stable under compounding shift severity.


\textbf{Performance on Source Domain.}
A key concern in continual test-time adaptation is \emph{forgetting}: gains under shifted conditions should not degrade behavior when the environment returns to nominal. 
To evaluate this, we append the source domain (flat terrain without domain shift) at the end of the continual sequence and directly assess the adapted policies (Appendix~\ref{sec:source_domain}, Fig.~\ref{fig:continual_then_source}). 
We observe that both \textsc{TEMPO} and \textsc{QMAX} recover source-domain performance \emph{instantly}, matching the original policy without any transient degradation or additional adaptation, indicating no observable forgetting.



\subsection{Ablating Components of TEMPO}

Table~\ref{tab:tempo_components} ablates the two components of \textsc{TEMPO}—entropy regularization and the score-based memory—across increasing severities of payload and terrain shifts. 
Starting from \textsc{QMAX}, adding \emph{entropy regularization} yields consistent gains of $+0.5$ to $+1.6$ return across the six settings. 
Replacing FIFO with the \emph{score-based memory} provides larger improvements of $+1.0$ to $+2.3$. Combining both (\textsc{TEMPO}) gives the best results in every column, improving over \textsc{QMAX} by $+1.4$ to $+2.5$.
This indicates that the design of the memory can play a crucial role in boosting the adaptation performance.
Relative to \textsc{No Adapt}, \textsc{TEMPO} improves by $+2.6$ to $+4.9$ per setting, indicating that both components contribute and their combination is complementary.

\begin{table}[t]
\centering
\caption{\textbf{Ablating \textsc{TEMPO}.}
Average episodic return under increasing shift severity for three corruption families.
Adding entropy regularization and score-based memory each improves over \textsc{QMAX}, and combining both (\textsc{TEMPO}) yields the best robustness across all severities.}
\begin{tabular}{lcccccc}
\toprule
Corruption &
\multicolumn{2}{c}{Mass on Calf (kg)} &
\multicolumn{2}{c}{Mass on Torso (kg)} &
\multicolumn{2}{c}{Rocky Terrain (cm)} \\
\cmidrule(lr){0-0}
\cmidrule(lr){2-3}
\cmidrule(lr){4-5}
\cmidrule(lr){6-7}
\multicolumn{1}{c}{\diagbox{Method}{Severity}}& 4 & 5 & 5 & 7.5 & 7.5 & 10 \\
\midrule
No Adapt          & 15.3 & 13.0 & 17.5 & 12.7 & 14.0 &7.0  \\
\midrule
QMAX              &  16.5& 14.4 & 19.7  & 15.4 & 16.6 &9.0  \\
\midrule
\,\,+ Entropy Max.    &  17.0&  15.3& 21.0 &  17.0& 17.3 & 10.6 \\
\,\,+ Score-Based Mem. &  17.5&  15.9  &  21.0 &  17.1& 17.8& 11.3  \\
\midrule
\,\,+ ALL (TEMPO)            & \textbf{17.9} &  \textbf{16.0}&  \textbf{21.5}& \textbf{ 17.6}&  \textbf{18.0} & \textbf{11.5}  \\
\bottomrule
\end{tabular}\label{tab:tempo_components}
\end{table}



\subsection{RMA as an Adaptation-Aware Backbone}
\label{sec:rma}

\begin{wrapfigure}{r}{0.51\columnwidth}  
  \centering
  \vspace{-20pt}
  \includegraphics[width=\linewidth]{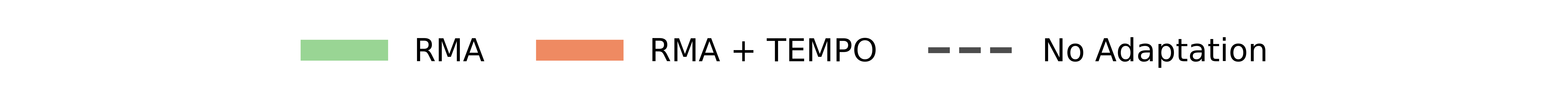}\par
  \vspace{2pt}
  \includegraphics[width=0.49\linewidth]{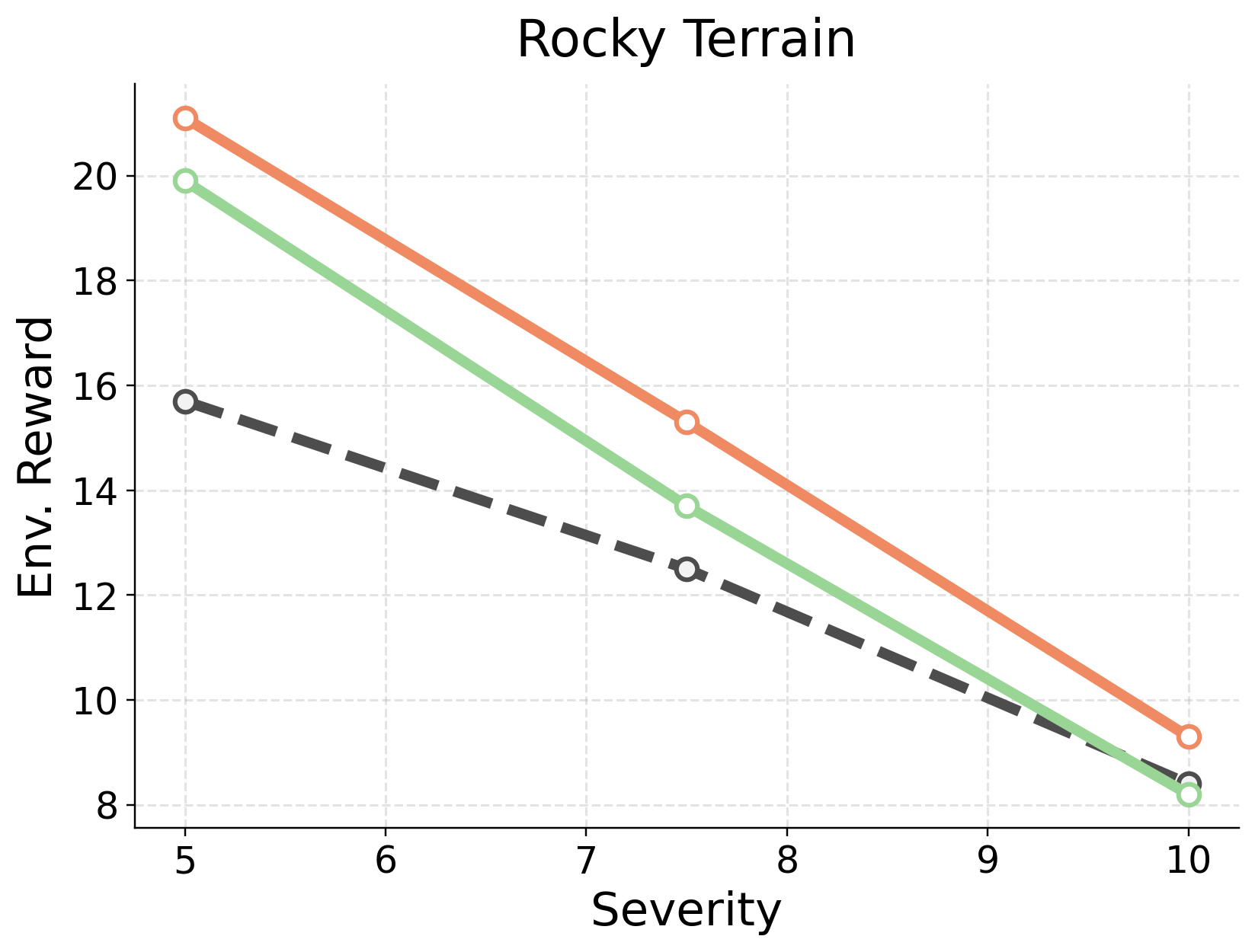}\hfill
  \includegraphics[width=0.49\linewidth]{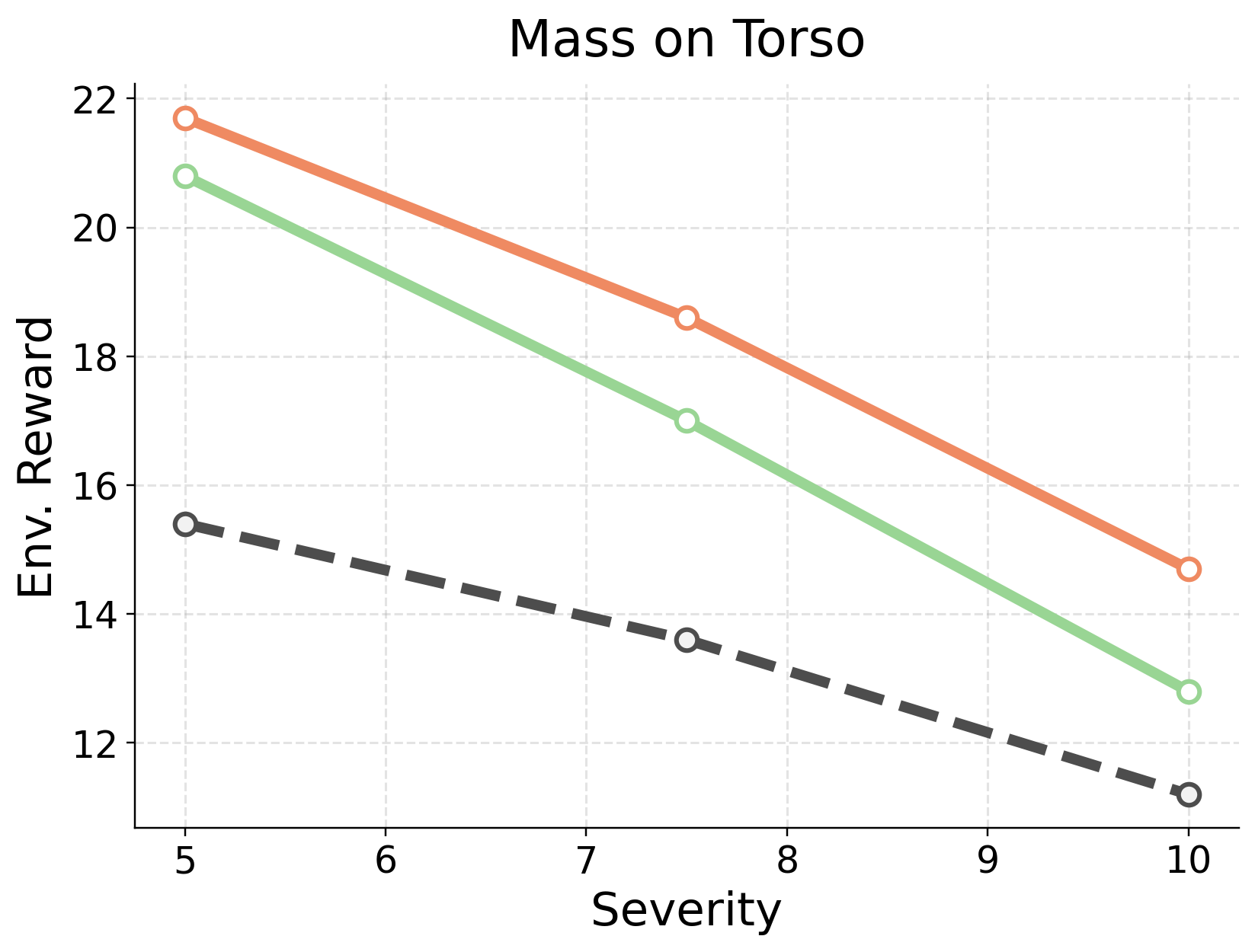}
  \vspace{-6pt}
  \caption{\textbf{RMA ablation.} Frozen RMA vs native RMA adaptation vs RMA+TEMPO. TEMPO provides consistent and significant additional gains.}
  \label{fig:rma}
\end{wrapfigure}

Rapid Motor Adaptation (RMA)~\citep{kumar2021rma} explicitly targets fast adaptation to dynamics
changes by conditioning the policy on a learned latent embedding that is inferred online during
deployment. Unlike the policies considered in previous sections, RMA is therefore not a static
pretrained controller, but already incorporates an internal adaptation mechanism, making it a strong
test for whether our proposed TEMPO can provide additional gains beyond architecture-specific
adaptation.

We evaluate three variants under the same shifted test streams:
\emph{(i) No Adaptation} (freeze RMA at deployment), \emph{(ii) RMA} (enable RMA’s native online
adaptation), and \emph{(iii) RMA + TEMPO} (apply TEMPO’s critic-guided test-time updates on top of
the deployed RMA policy).
Figure~\ref{fig:rma} shows that native RMA adaptation improves over the frozen
baseline, while augmenting RMA with TEMPO yields the best performance across the evaluated shift
severities, demonstrating that TEMPO complements (rather than replaces) RMA’s built-in adaptation.
Further experiments on different architectures, and robustness analyses of TEMPO can be found in  Appendix~\ref{sec:app_exps}.




\section{Conclusions}

In this work, we studied reward-free, on-device test-time adaptation for robot locomotion under deployment-time distribution shift. Starting from a lightweight baseline that updates only the actor using a frozen pretrained critic, we introduced TEMPO, which stabilizes critic-guided adaptation through entropy regularization and a score-based memory that prioritizes informative test-time observations. 
Across simulated locomotion benchmarks and multiple shift types, architectures, and environments, TEMPO consistently improves over no adaptation and naive critic maximization, with the clearest gains under severe shifts. 
We further confirm the same trend on a physical Unitree Go2 via fully on-device adaptation under a calf-payload perturbation.

{
\small
\bibliographystyle{unsrtnat}
\bibliography{main}
}


\clearpage
\appendix


\section{Test-time Entropy regularized, Memory guided Policy Optimization}\label{sec:app_algos}
\begin{figure*}[b]
\vspace{-2mm}
\centering

\begin{minipage}[t]{0.49\textwidth}
\captionsetup{type=algorithm}
\hrule height 1.1pt
\captionof{algorithm}{\textsc{TEMPO} (high-level online loop)}
\label{alg:tempo_short}
\vspace{-2mm}

\hrule height 0.8pt
\vspace{0.8ex}

{\footnotesize
\textbf{Input:} pretrained $\pi_{\theta_0}$; frozen critic $Q_{\psi_0}$; 
step size $\eta$.\\
\textbf{Output:} actions $\{a_t\}_{t\ge 0}$ and adapted parameters $\{\theta_t\}_{t\ge 0}$.
}

\vspace{0.6ex}
\hrule height 0.4pt
\vspace{0.6ex}

\begin{algorithmic}[1]
\STATE Initialize $\theta_0$, $\mathcal{B}\leftarrow\emptyset$
\FOR{$t=0,1,2,\dots$}
  \STATE  Observe $o_t \sim \hat P(o_{t-1},a_{t-1})$
  \STATE  Act $a_t \leftarrow \pi_{\theta_t}(o_t)$
  \STATE  $\mathcal{B} \leftarrow \textsc{UpdateMemory}(\mathcal{B}, o_t)$
  \IF{\textsc{ShouldAdapt}$(t,\mathcal{B})$}
    \STATE $\theta_{t+1} \leftarrow \text{TempoUpdate}(\theta_t, \mathcal B, Q_{\psi_0})$
  \ELSE
    \STATE  $\theta_{t+1} \leftarrow \theta_t$
  \ENDIF
\ENDFOR
\end{algorithmic}
\label{alg:tempo}
\vspace{0.6ex}
\hrule height 0.8pt

\vspace{0.8ex}
{\footnotesize
\textbf{\textsc{ShouldAdapt}}: true every $K$ steps when $|\mathcal{B}|\ge B$.
}
\end{minipage}
\hfill
\begin{minipage}[t]{0.49\textwidth}
\captionsetup{type=algorithm}
\hrule height 1.1pt
\captionof{algorithm}{UpdateMemory (Score Based)}
\label{alg:score_memory}
\vspace{-2mm}
\hrule height 0.8pt
\vspace{0.6ex}

{\footnotesize
\textbf{Input:} observation $o$; memory bank $\mathcal B$; 
\\ \textbf{Output:} updated memory bank $\mathcal B$
}

\vspace{0.6ex}
\hrule height 0.4pt
\vspace{0.6ex}

\begin{algorithmic}[1]
\STATE \hspace{0.8em} $a \leftarrow \pi_{\theta}(o)$
\STATE \hspace{0.8em} $S \leftarrow \frac{1}{Q_{\psi_0}(o,a)+\varepsilon}$ \hfill \COMMENT{Eq.~\ref{eq:score}}
\IF{$|\mathcal{B}| < N$}
  \STATE \hspace{0.8em} Insert $(o,S)$ into $\mathcal{B}$
\ELSE
  \STATE \hspace{0.8em} $(o_{\min},S_{\min}) \leftarrow \arg\min_{(o',S')\in\mathcal{B}} S'$
  \IF{$S > S_{\min}$}
    \STATE \hspace{0.8em} Remove $(o_{\min},S_{\min})$ from $\mathcal{B}$
    \STATE \hspace{0.8em} Insert $(o,S)$ into $\mathcal{B}$
  \ENDIF
\ENDIF
\end{algorithmic}

\vspace{0.6ex}
\hrule height 0.8pt
{\footnotesize
\vspace{0.8ex}
\textbf{\textsc{TempoUpdate}}: Sample with Eq.\ref{eq:importance_sampling} and optimize Eq~\ref{eq:tempo_ent}.
}
\end{minipage}
\hfill
\vspace{-2mm}\end{figure*}

In Section~\ref{sec:our_method}, we proposed TEMPO, an efficient, on-device friendly TTA method for adapting robotic policies. 
In the construction of TEMPO, we followed~\cite{seo2025fasttd3} in accessing the distributional critic. 
We note that the output of the distributional critic is nothing but the output of the penultimate layer of the critic network.
Refer to the works of ~\cite{seo2025fasttd3, lee2025simbav2} for further details.
Due to the space constraints, we did not provide a full explicit algorithm to describe how one can implement TEMPO in practice.
For completeness, we provide these algorithms in this Section.

Algorithm~\ref{alg:tempo} summarizes the high level interaction between the test stream and the pretrained policy equipped with our proposed TEMPO.
Given an observation vector $o_t$, the policy decides on the action $a_t$. 
TEMPO leverages this observation vector and its corresponding action in updating its memory bank $\mathcal B$ according to Algorithm~\ref{alg:score_memory}.
When there are enough samples in the memory to conduct an adaptation step, TEMPO samples a batch with importance sampling according to Eq.~\ref{eq:importance_sampling} and runs a single update step with the loss in Eq~\ref{eq:tempo_ent}.
We note that for efficiency purposes, TEMPO conducts a single update step every $K$ received observations to keep the additional compute requirement light.

Algorithm~\ref{alg:score_memory} describes the score based memory update that we proposed in section~\ref{sec:score_memory}.
In a nutshell, as long as the memory bank not full, we store all received observation accompanied by their corresponding score.
When hitting the capacity limit, we check if the newly received observation has a better score than the sample with the lowest score in memory, then we evict that sample and store the newly received one.
We note that for additional implementation efficiency purposes, one can leverage directly the action from step 4 in Algorithm~\ref{alg:tempo} instead of recomputing it in step 1 in Algorithm~\ref{alg:score_memory} (one less forward pass).

\section{Experiments}\label{sec:app_exps}

In Section~\ref{sec:experiments}, we provided comprehensive evaluation of how TEMPO is effective under different scenarios in both simulation and real world setting.
In this section, we provide additional details about the experimental setup for full reproducibility, and additional analysis on TEMPO.

\subsection{Setup.}\label{sec:app_exps_setup}

The policy observation consists of proprioceptive and command-related signals commonly used in legged locomotion. Specifically, it includes IMU measurements in the form of angular velocity (gyro) and a gravity-aligned vector, base linear velocity, joint angles expressed relative to a default pose, joint velocities, the previous action, and a commanded motion vector $(v_x, v_y, w_z)$ specifying desired forward/lateral linear velocities and yaw rate.
The critic uses a symmetric, non-privileged observation space identical to the policy input, which enables on-device test-time adaptation directly from robot sensors. Since our real-world Go2 robot does not enable linear velocity estimation out of the box, we utilize an Extended Kalman Filter to calculate velocities on the fly using the built-in foot force sensors \citep{hartley2018contact}.

\subsection{Generalization to Different Architectures and Environments  }\label{sec:domain_randomization}

A central goal of our approach is \emph{portability}: test-time adaptation should improve a deployed controller without relying on architecture-specific components, privileged signals, or task-tailored objectives. 
We therefore evaluate whether \textsc{TEMPO} transfers beyond our default setting (Fast-TD3 on Go2) to substantially different pretrained policy families and to a different embodiment, without modifying the algorithm or introducing additional modules.

\begin{figure}[t]
    \centering
    \includegraphics[width=0.5\linewidth]{figures/shared_legend/legend_bars.png}
    \\
    \includegraphics[width=0.32\linewidth]{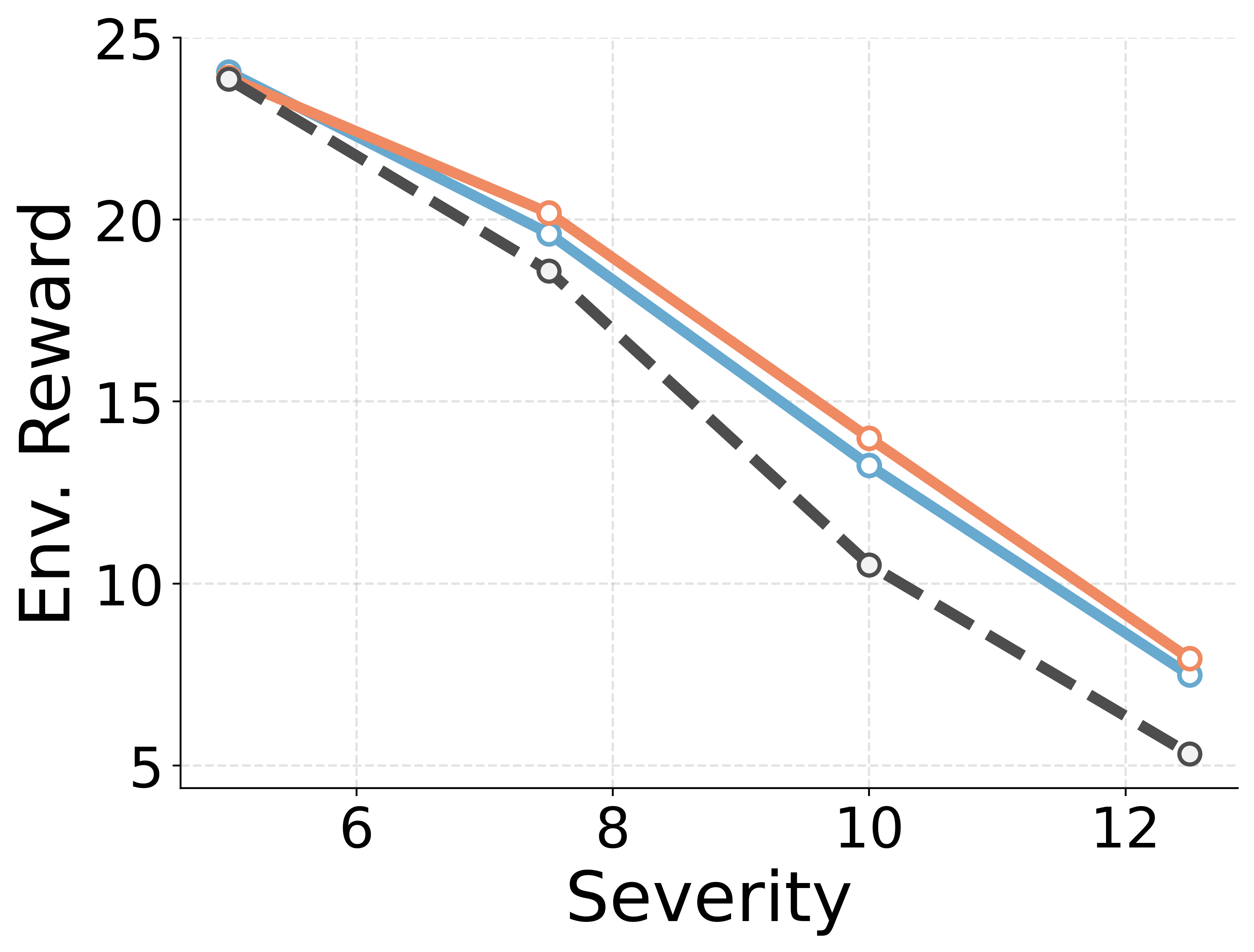}
    \includegraphics[width=0.32\linewidth]{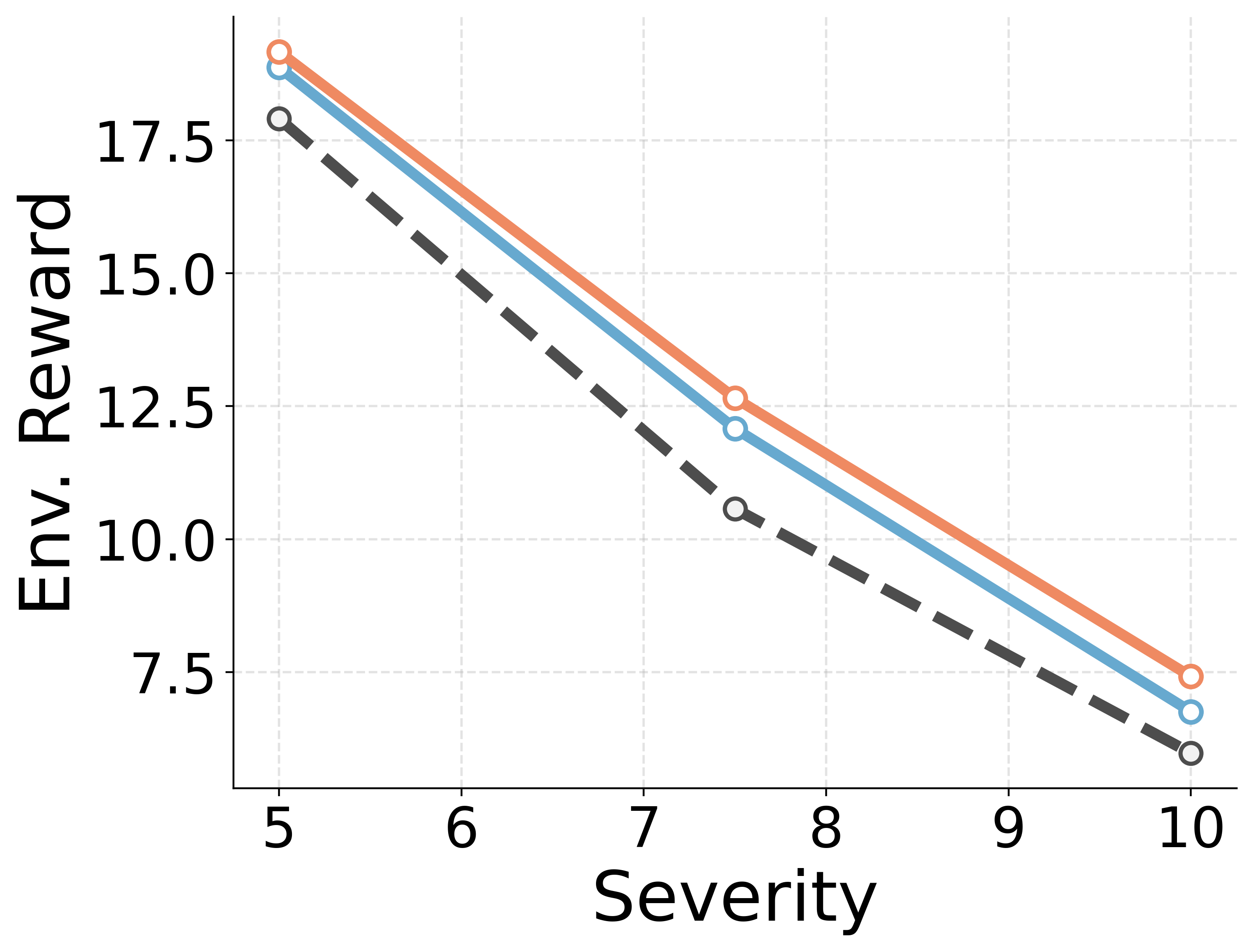}
    \includegraphics[width=0.32\linewidth]{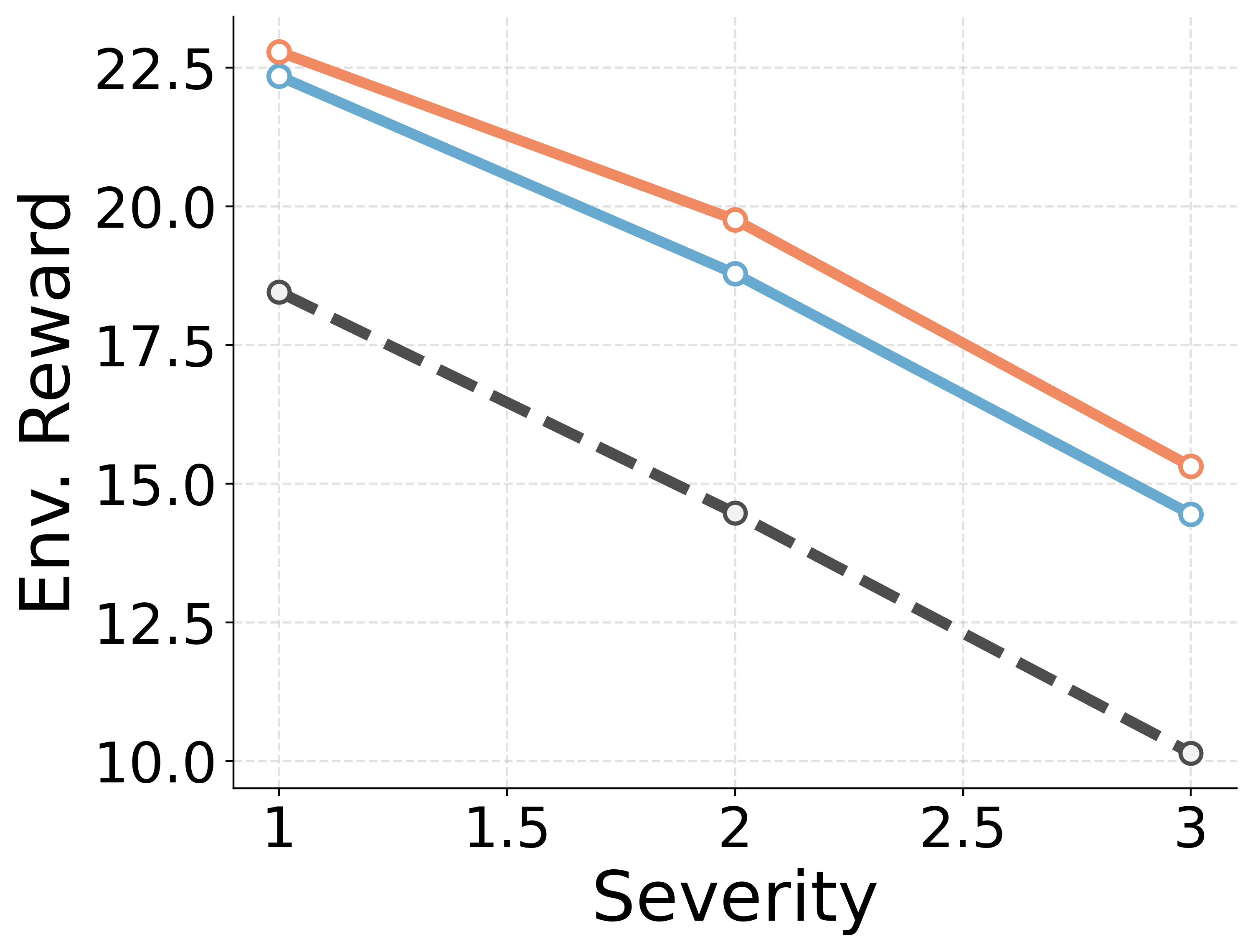}
    \caption{\textbf{Generalization across architectures and embodiments.}
    Rocky-terrain results (left$\rightarrow$right): Fast-TD3 trained with domain randomization, SimBA-V2, and the G1 humanoid. \textsc{TEMPO} transfers without architecture-specific modifications and remains effective across settings.}
    \label{fig:different_archs}
\end{figure}

Figure~\ref{fig:different_archs} summarizes this evaluation on the \emph{rocky terrain} test environment. On the left, we consider a Fast-TD3 policy trained \emph{with domain randomization}, representing a stronger robustness-oriented pretraining baseline. 
Table~\ref{tab:additional_embodiments} reports results on Go1 and G1 under mass perturbations.
Even in this regime, \textsc{TEMPO} continues to provide consistent improvements, indicating that its gains are not limited to policies trained without robustness regularization and that it can complement domain randomization at deployment. 
In the middle, we adapt a policy trained with the \textsc{SimBA-V2} architecture, which differs in inductive biases and learning dynamics from Fast-TD3; \textsc{TEMPO} remains effective without any architecture-specific tuning, supporting the claim that our objective and update rule are broadly applicable to actor--critic policies. 
Finally, on the right, we evaluate on \textsc{G1}, a humanoid with markedly different morphology and contact dynamics. Despite this shift in embodiment, \textsc{TEMPO} still improves performance on rocky terrain, suggesting that the method generalizes beyond quadrupeds and can operate in more challenging control regimes.

\begin{table}[b]
\centering\vspace{-0.5cm}
\caption{Additional results on Go1 and G1 under payload perturbations. MC denotes mass on calf and MT denotes mass on torso.}
\label{tab:additional_embodiments}
\begin{tabular}{lccccccc}
\toprule
& \multicolumn{4}{c}{Go1} & \multicolumn{3}{c}{G1} \\
\cmidrule(lr){2-5}
\cmidrule(lr){6-8}
Method & MC=4 & MC=5 & MT=7.5 & MT=10 & MT=2 & MT=3 & MT=5 \\
\midrule
No Adapt & 18.9 & 16.3 & 13.6 & 6.8 & 21.3 & 17.5 & 10.0 \\
QMAX & 19.6 & 17.6 & 13.6 & 6.9 & 24.1 & 22.1 & 17.4 \\
TEMPO & \textbf{20.5} & \textbf{18.6} & \textbf{16.9} & \textbf{11.0}
      & \textbf{24.5} & \textbf{23.4} & \textbf{18.5} \\
\bottomrule
\end{tabular}
\end{table}

\subsection{Sensitivity to Hyperparameters, Memory Size, Batch Size, and Adaptation Interval  }\label{sec:memory_size}

\begin{figure}[t]
    \centering
    \includegraphics[width=0.5\linewidth]{figures/shared_legend/legend_bars.png}
    \\
    \includegraphics[width=0.32\linewidth, trim=0 0 0 1.5cm,clip]{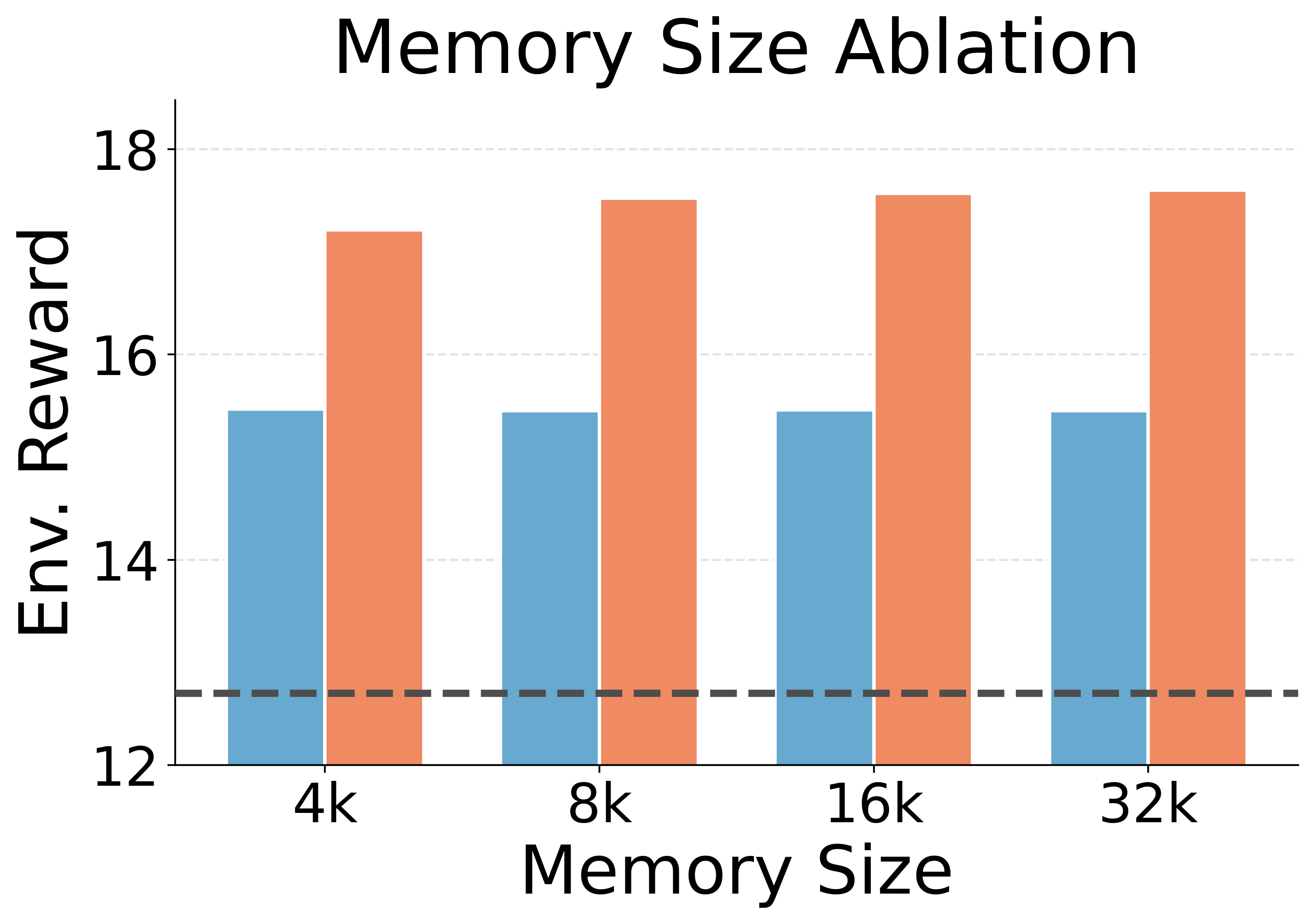}
    \includegraphics[width=0.32\linewidth, trim=0 0 0 1.5cm,clip]{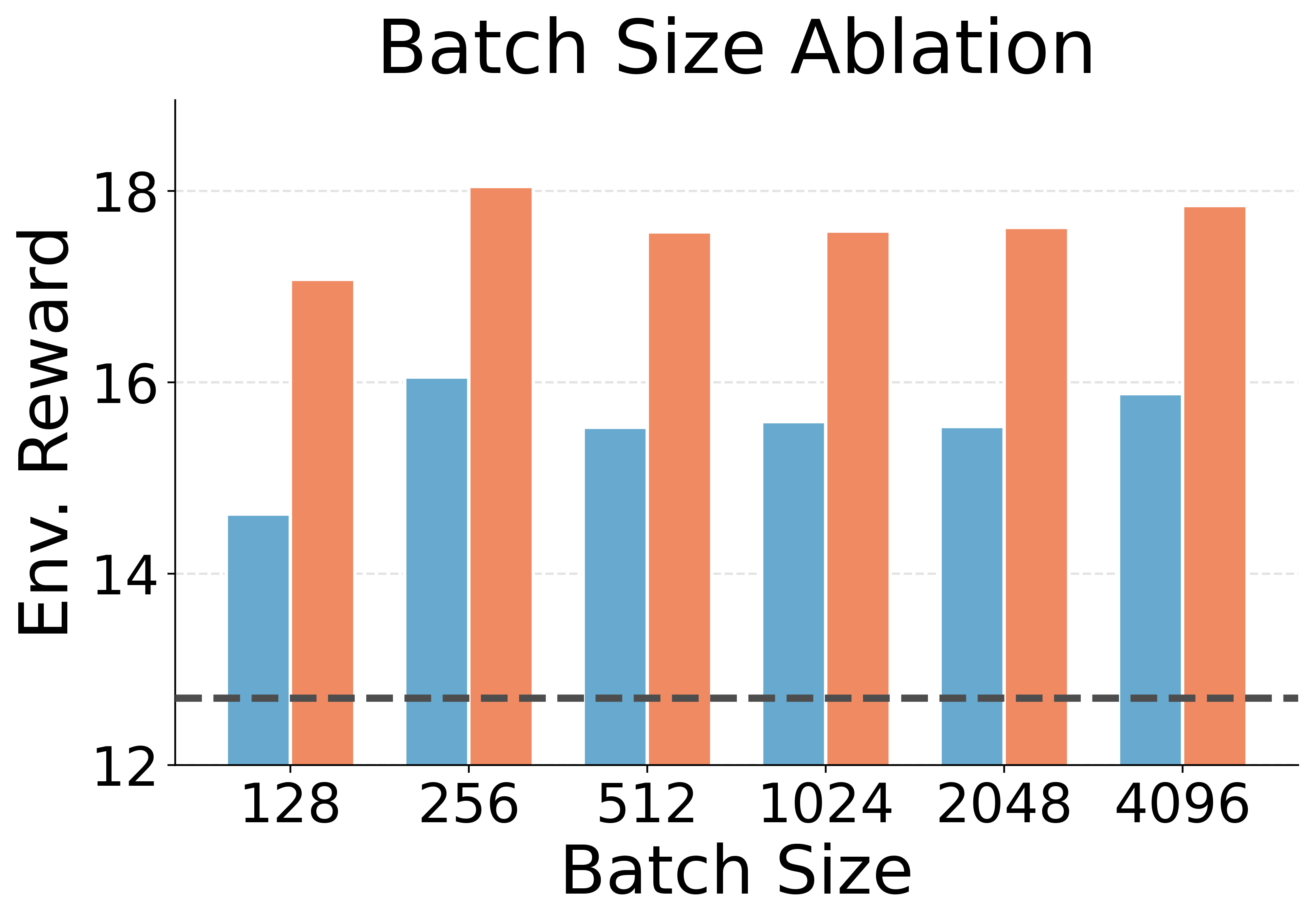}
    \includegraphics[width=0.32\linewidth, trim=0 0 0 1.5cm,clip]{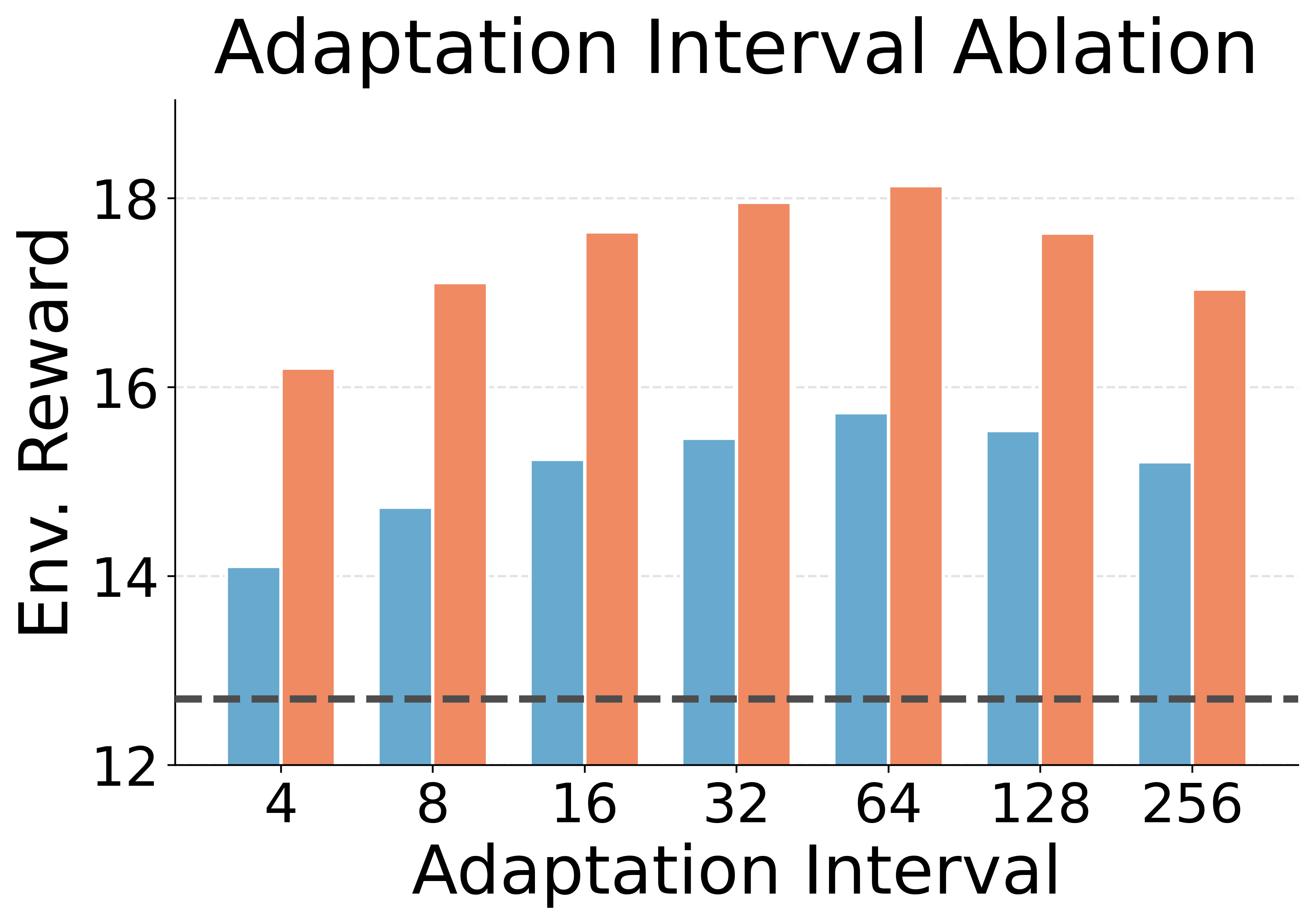}
    \caption{\textbf{Robustness to on-device adaptation hyperparameters.} \textsc{TEMPO} is robust and performs best across different memory sizes (left), batch sizes (middle), and adaptation intervals (right).
    All experiments are run with Fast-TD3 on Go2 with Mass on Torso Corruption with severity of 7.5 kg.}
    \label{fig:ablations-sensitivity}
\end{figure}

On-device test-time adaptation must operate under strict memory and compute budgets; accordingly, we study how \textsc{TEMPO} behaves as we vary the three deployment-critical hyperparameters that directly control resource usage: replay memory size $|\mathcal{B}|$ (storage footprint), minibatch size $B$ (compute per update), and adaptation interval $K$ (update frequency, i.e., one gradient step every $K$ environment steps). Figure~\ref{fig:ablations-sensitivity} reports performance under distribution shift across these ablations.

Overall, \textsc{TEMPO} is robust across a broad range of settings and remains the best-performing method throughout the three sweeps. Importantly, this robustness translates into practical efficiency on device. First, \textsc{TEMPO} maintains strong performance even with relatively small $|\mathcal{B}|$, implying that effective adaptation does not require storing large histories; this keeps the on-device memory footprint small and compatible with embedded constraints. Second, \textsc{TEMPO} performs well with small minibatches, showing that adaptation can be carried out with limited per-update compute (and correspondingly lower latency), rather than relying on large, expensive updates. Third, \textsc{TEMPO} remains effective for larger adaptation intervals $K$, meaning updates can be applied infrequently—reducing the total number of backward passes per unit time and making the method applicable even when compute budgets are very tight.

As expected, extreme settings reveal intuitive trade-offs. Very small memories reduce state diversity and can limit coverage of informative ``hard'' states, while very large memories yield diminishing returns given a fixed update schedule. Similarly, very small batches increase gradient noise, whereas very large batches offer limited additional benefit relative to their higher compute cost. Finally, $K$ exposes a responsiveness--efficiency trade-off: smaller $K$ increases adaptation frequency but raises compute load, while larger $K$ lowers overhead at the cost of slower reaction to abrupt changes. Taken together, the results indicate that \textsc{TEMPO} provides stable improvements without delicate tuning and can be configured to meet stringent on-device constraints via small $|\mathcal{B}|$, small $B$, and large $K$.

\begin{wrapfigure}{r}{0.27\columnwidth}  
  \centering
  \vspace{-5pt}
  \vspace{2pt}
  \includegraphics[width=0.9\linewidth, trim=0 0 0 1.4cm,clip]{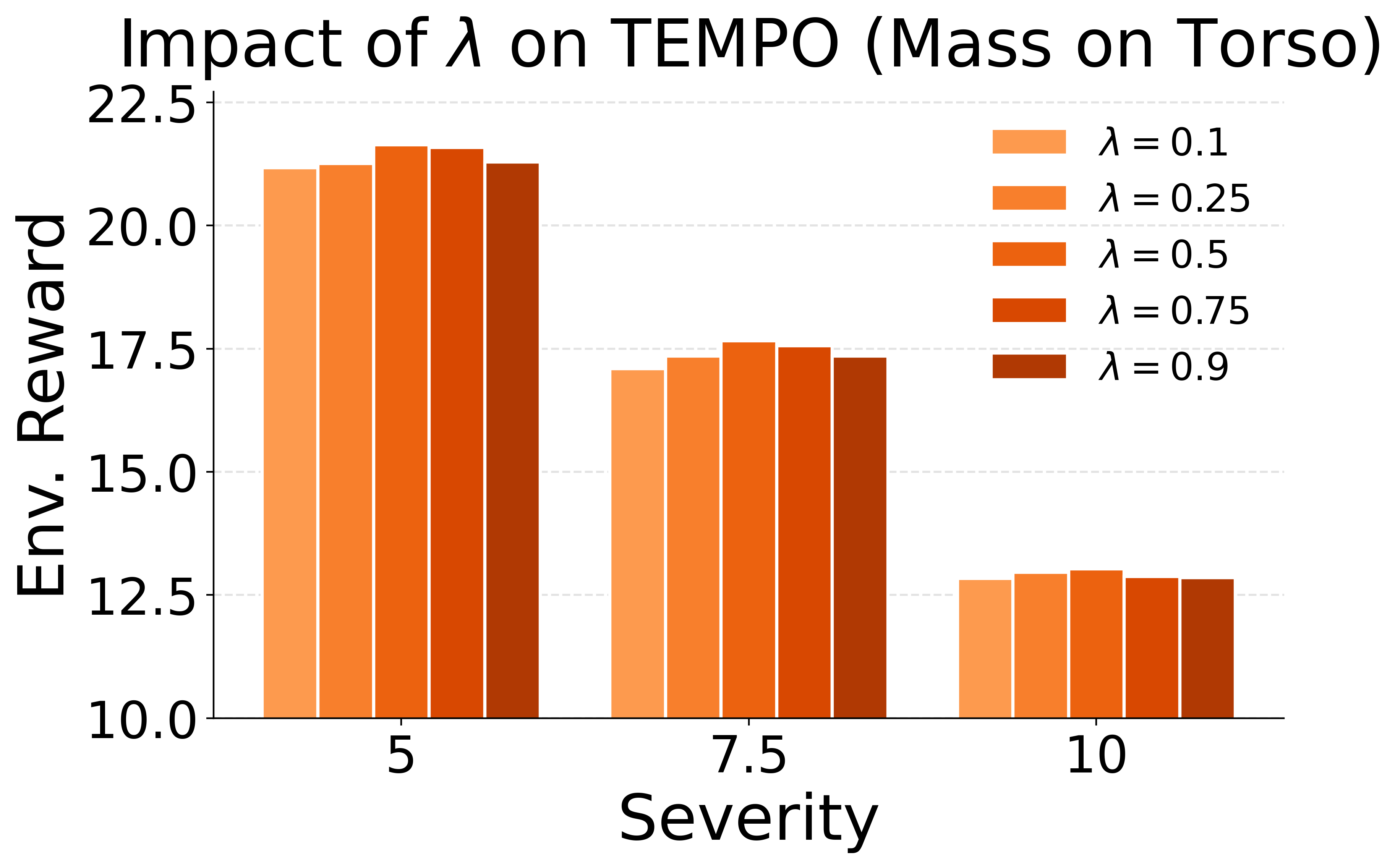}
  \vspace{-6pt}
  \caption{\textbf{Ablating $\lambda$.}}\vspace{-16pt}
  \label{fig:lambda}
\end{wrapfigure}

At last, we assess the performance of TEMPO when varying its only hyperparameter; $\lambda$. We follow the experimental setup in Figure~\ref{fig:ablations-sensitivity} and vary $\lambda \in \{0.1, 0.25, 0.5, 0.75, 0.9\}$ and report the results in Figure~\ref{fig:lambda}. We observe a stable behavior under different values of $\lambda$.

\subsection{Performance on Source Domain} \label{sec:source_domain}

A key concern in continual test-time adaptation is catastrophic forgetting: parameters adapted to severe deployment shifts may degrade performance once the environment returns to nominal conditions. This risk is particularly acute in on-device, online settings, where updates are applied continuously without explicit supervision, replay from the source domain, or reset mechanisms.
To assess whether critic-guided test-time adaptation induces such forgetting, we evaluate both QMAX and TEMPO after prolonged exposure to increasingly severe distribution shifts, followed by a return to the source domain (flat terrain without perturbations). Specifically, we adopt the continual TTA protocol from Section~5.4 and append a final evaluation phase in which the environment dynamics match the original training conditions, while keeping the adapted policy parameters fixed—without additional fine-tuning or recovery steps.

\begin{figure}[t]
    \centering
    \includegraphics[width=0.75\linewidth]{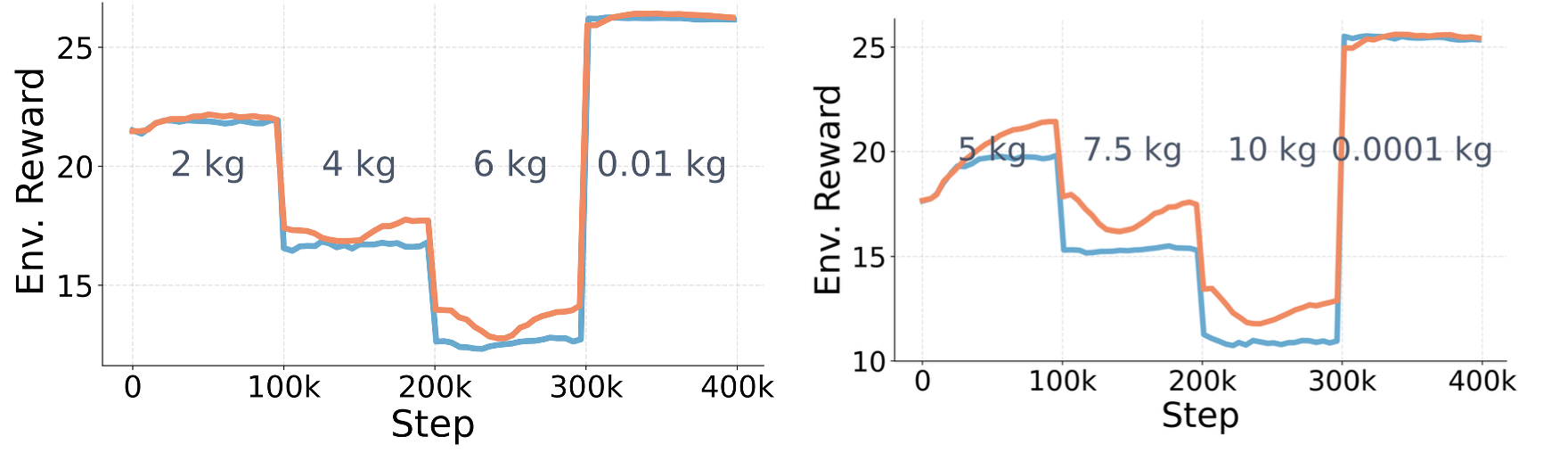}
    \caption{\textbf{Continual TTA followed by return to the source domain.} Performance under a non-stationary deployment stream in which the severity of the domain shift increases over time and is then reset to nominal conditions. Left: mass-on-calf shift. Right: mass-on-torso shift. Architecture: Fast‑TD3. Both QMAX and TEMPO recover source-domain performance immediately after the shift is removed, indicating no catastrophic forgetting despite prolonged adaptation under shift.}
    \label{fig:continual_then_source}
\end{figure}

Figure~\ref{fig:continual_then_source} reports performance for two representative shift families (mass on calf and mass on torso) using the Fast-TD3 architecture. In both cases, QMAX and TEMPO immediately recover source-domain performance, matching the non-adapted policy without any transient degradation. This indicates that critic-guided test-time updates do not overwrite nominal behaviors, and that adaptation remains localized to state regions associated with deployment-time shifts.
Overall, these results suggest that both methods—particularly TEMPO—support stable continual adaptation without observable forgetting, making them well suited for long-horizon deployment scenarios in which operating conditions drift temporarily but may later revert to nominal regimes.

Further, and to complement the previous experiments, We evaluate the trained policy under the source domain (flat terrain, no shift) with no adaptation, QMAX, and TEMPO, across all three reward signals used in the paper (averaged across six seeds). We observe that all three methods are numerically indistinguishable, and both velocity-tracking signals sit within run-to-run noise (environment reward of 25.6, Linear Velocity tracking reward of 864, and Angular Velocity Tracking reward of 457).

\subsection{Improving Angular Velocity Tracking} \label{sec:other_rewards}
While the main paper focuses on task-level performance and forward locomotion metrics, stable angular velocity tracking is critical for practical deployment, especially under distribution shifts that alter contact dynamics or inertial properties. 
Poor tracking of angular velocity often manifests as oscillatory turning, delayed heading response, or loss of stability, which may not be fully reflected in episodic return alone.
To examine this aspect more closely, we evaluate the angular velocity tracking performance of QMAX and TEMPO on Go2 under the same deployment-time distribution shifts considered in Section~\ref{sec:main_results} For each method, we measure the tracking angular velocity reward, and report results averaged across 6 random seeds.

\begin{figure}[b]
    \centering
    \hspace{2cm}\includegraphics[width=0.4\linewidth]{figures/shared_legend/legend_bars.png}
    \newline
    \includegraphics[width=0.245\linewidth]{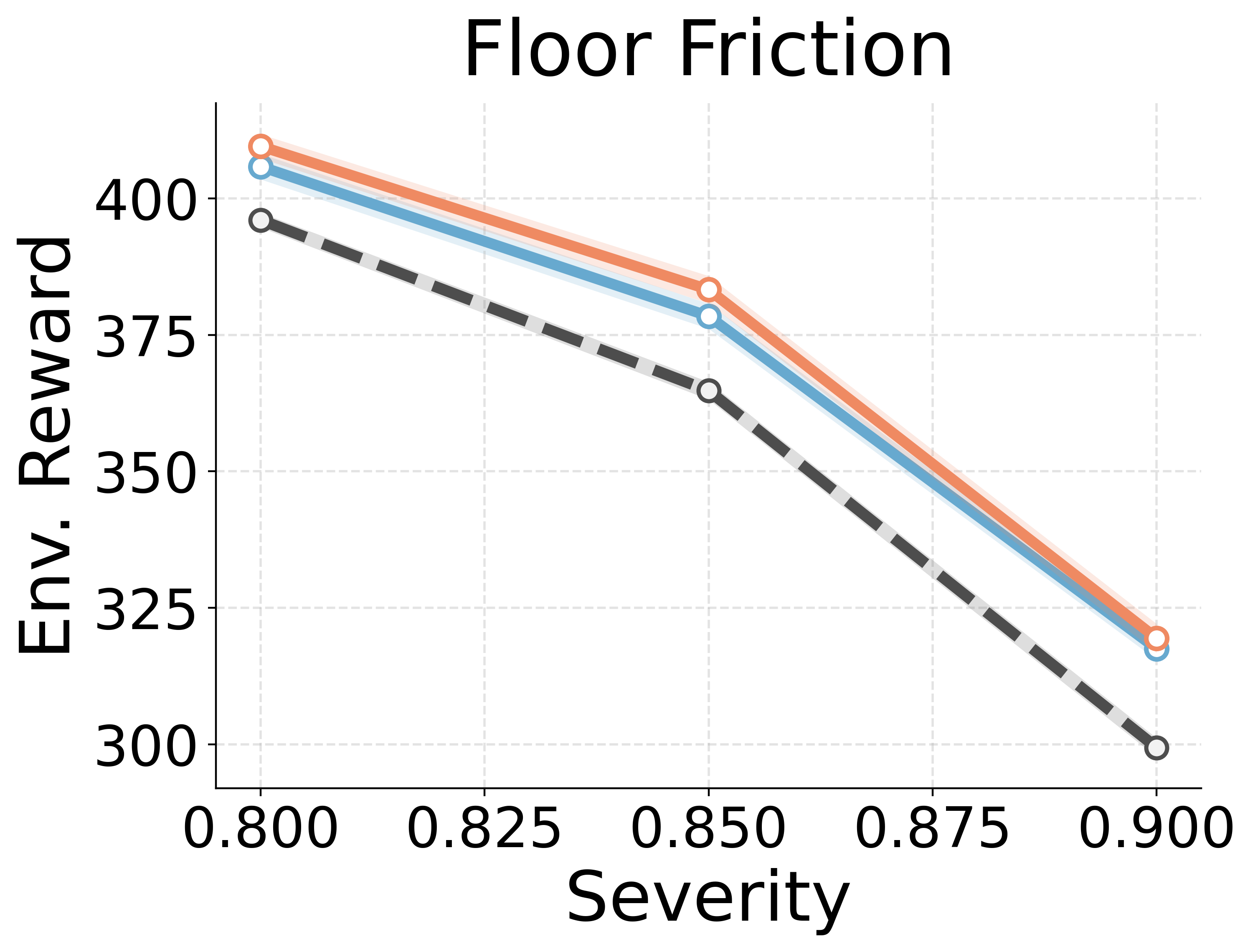}
    \includegraphics[width=0.245\linewidth]{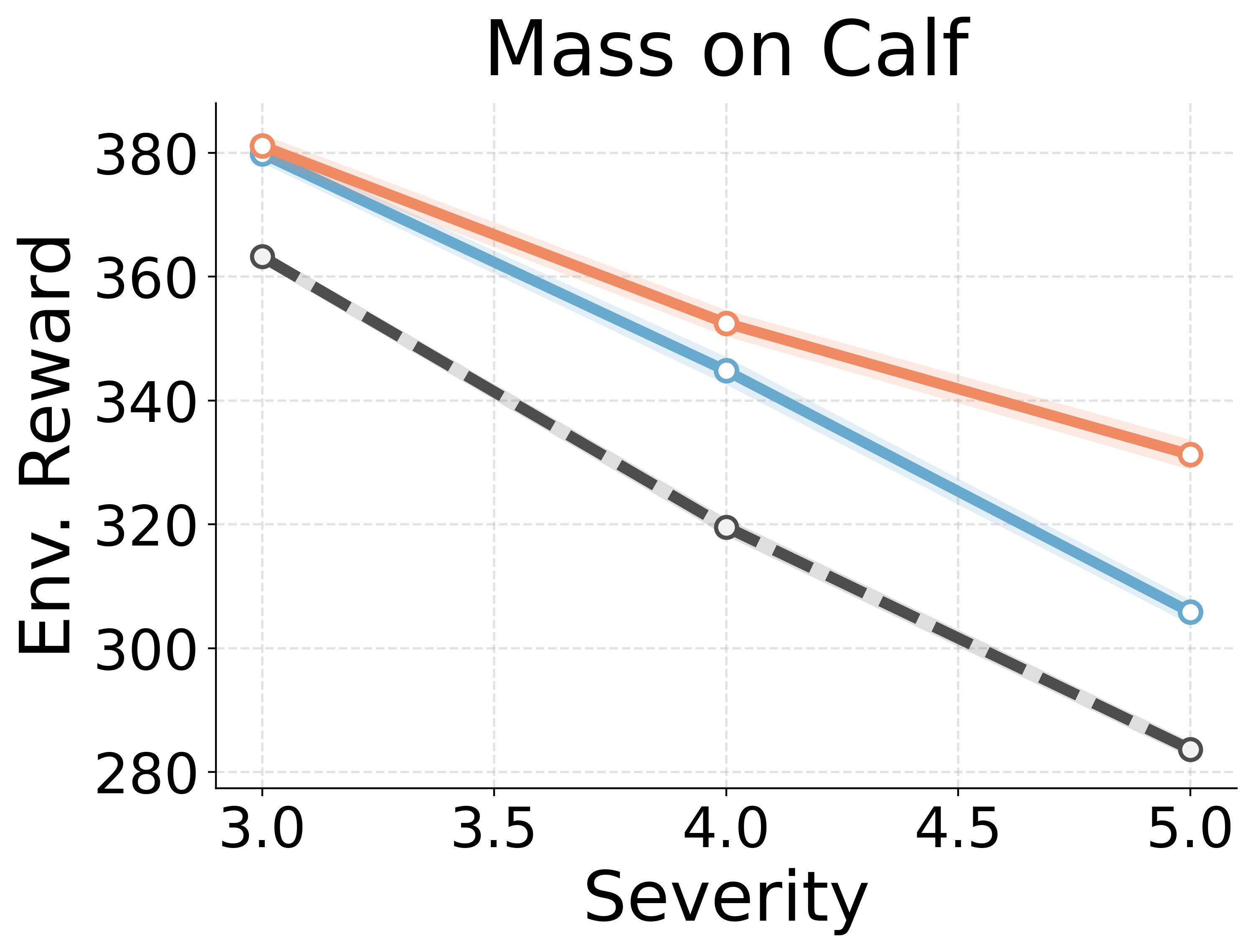}
    \includegraphics[width=0.245\linewidth]{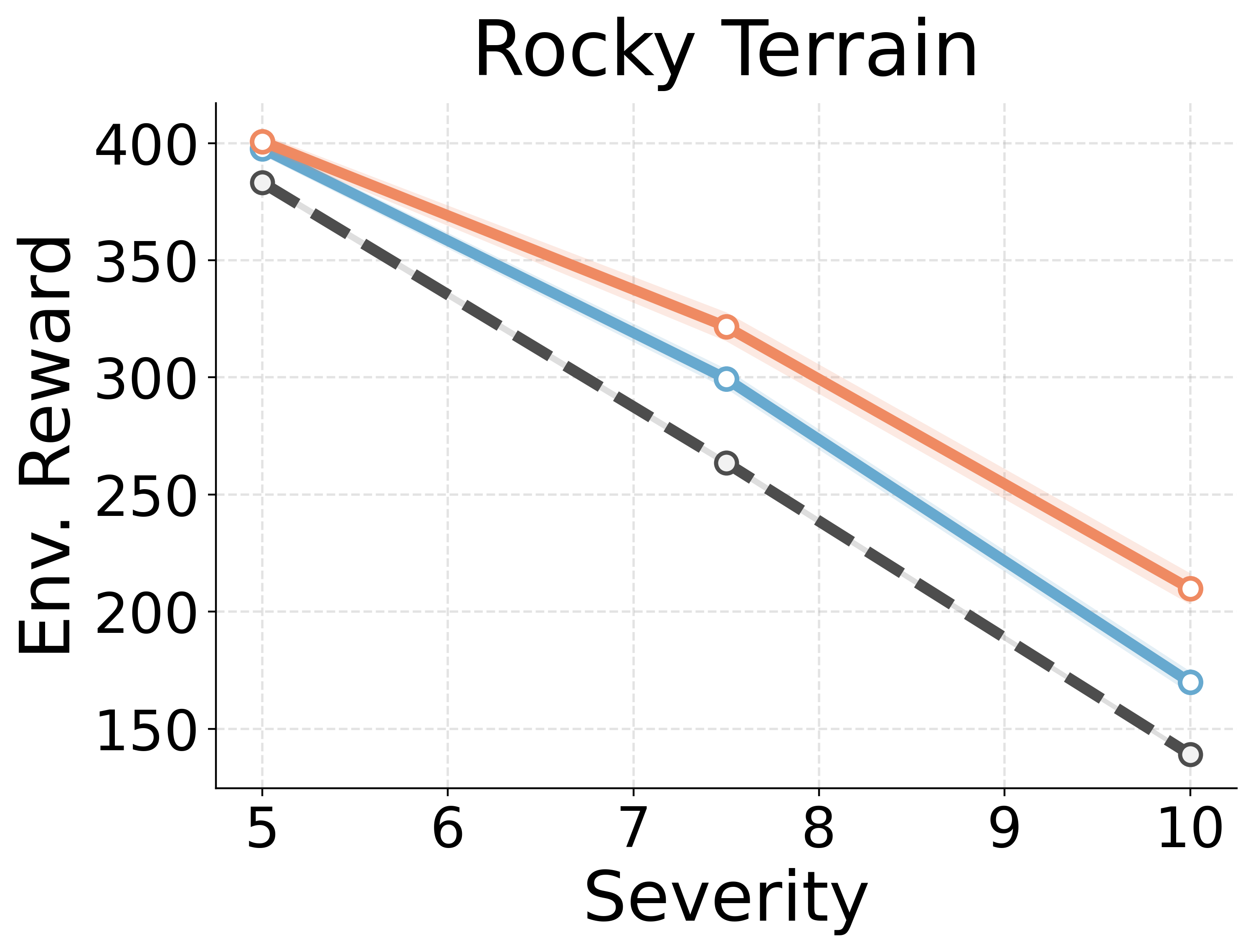}
    \includegraphics[width=0.245\linewidth]{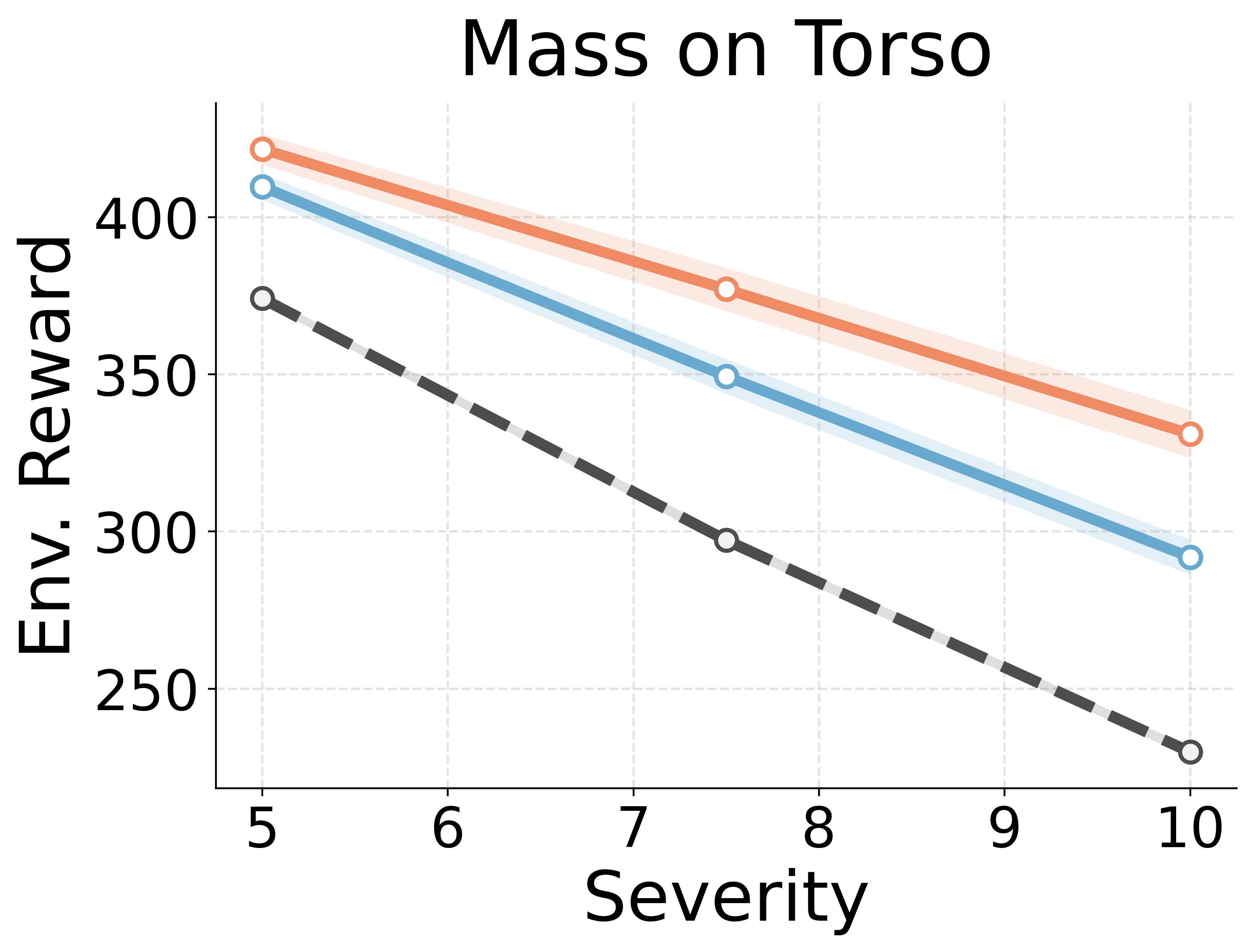}
    \caption{\textbf{Angular velocity tracking on Go2 under deployment-time distribution shift.} We report angular velocity tracking error as a function of shift severity for four corruption families. Results are averaged over five seeds. Critic-guided test-time adaptation improves tracking accuracy relative to no adaptation, with TEMPO achieving the most stable and accurate angular velocity tracking, particularly under severe distribution shifts.}
\label{fig:ang_vel}\vspace{-0.5cm}
\end{figure}

Figure~\ref{fig:ang_vel} summarizes the results across four shift families with increasing severity. 
Without adaptation, the pretrained policy exhibits rapidly degrading angular velocity tracking as the shift intensifies. 
Introducing critic-guided test-time adaptation substantially mitigates this degradation: QMAX consistently improves tracking accuracy relative to no adaptation, while TEMPO provides the most stable and accurate tracking across all severities. 
The gains are most pronounced under harder shifts, such as rocky terrain and torso mass perturbations, where overconfident critic-driven updates can otherwise lead to unstable turning behavior.
Overall, these results indicate that the benefits of TEMPO extend beyond forward velocity and reward metrics to finer-grained control objectives, demonstrating improved stability and control fidelity under deployment-time distribution shift.

\subsection{Does TEMPO Improve Local Action Coverage?}
To analyze the impact of TEMPO on the local action coverage, we measure the critic's distributional entropy over the course of adaptation. On the severe mixed-corruption setting (Mass on Torso=5, Rough Terrain=0.06; rightmost column of Table~\ref{tab:mixed_corruptions}), starting from an initial entropy of 1.15, QMAX drives the entropy of $Z_{\psi_0}(o, \pi_\theta(o)$ down to 1.05 while TEMPO raises it to 1.41 — a 34\% relative gap that is temporally aligned with the return gap (9.9 vs 14.0, +41\% relative). Elevated critic entropy preserves broader gradient support and thus wider local action coverage, exactly the mechanism the regularizer is designed to induce. This is also consistent with our results in Table~\ref{tab:tempo_components}, where removing the entropy term reduces performance. 

\subsection{RMA Setup}
Due to the space constraints, we breifly outlined the experimental setup of RMA in Section~\ref{sec:rma}. Here we provide the remaining training details for full reproducibility.
RMA policy is trained on Go2JoystickFlatTerrain with MuJoCo Playground domain randomization enabled. The training randomization covers floor friction U(0.4,1.0), joint friction scale U(0.9,1.1), armature scale U(1.0,1.05), center-of-mass shift U(-0.05,0.05), link-mass scale U(0.9,1.1), torso mass offset U(-1,1) kg, and initial joint jitter U(-0.05,0.05). The RMA target conditions are deliberately outside these ranges: e.g., floor friction 0.1–0.2, torso payload 5–10 kg, calf payload 3–5 kg, and rough-terrain height 0.05–0.1, where terrain height is not part of the flat-terrain training randomization.

\subsection{Refreshing Scores of Memory}
Under long hoizon streams, scores in the memory bank for previously observed observations could become stale hindering the positive impact of adaptation.
A natural fix to this problem is to integrate refreshing the scores for the observations in the memory bank.
For that, we equipped TEMPO with either of the following two score refreshing variants: (i) periodic score refresh, and (ii) age-decayed scoring, where Eq. (3) is replaced by $\hat S(o) = S(o) / (1 + \gamma \cdot \mathrm{age}(o))$ with $\mathrm{age}(o)$ the number of steps $o$ has resided in the memory. We swept $\gamma \in {10^{-5}, 10^{-4}, 10^{-3}, 10^{-2}, 10^{-1}, 1, 10, 100}$. We observe that both variants added per-step compute without improving returns over the current minimum-score replacement rule. We attribute this to the limited state diversity of our simulation; in real-world, long-horizon deployment with richer diversity, we expect stale-score handling to matter more.

\subsection{Comparison to Test-Time Training}
Recently, \cite{park2024value} proposed OPEX and Test Time Training as adaptation methods for policies subjected to shifts. 
In Figure\ref{fig:pipeline}, we analyzed the efficacy of OPEX. For completeness, we discuss here the performance of TTT.
In particular, one can express the adaptation loss for TTT as:
$\mathcal{L}_{\mathrm{TTT}}=\mathcal{L}_{\mathrm{QMAX}}+\beta\,\mathrm{KL}\!\left(\pi_\theta \,\|\, \pi_{\theta_0}\right).$
Thus, we compare our proposed baseline; QMAX versus TTT. We evaluate on Go2 under the mass-on-calf shift and sweep the regularization coefficient $\beta$.

We observe a monotonic decrease in performance as $\beta$ increases. For MC=4, the return drops from $16.8$ at $\beta=0$ (QMAX) to $16.3$, $16.1$, $15.7$, and $15.6$ for $\beta=\{0.02,0.04,0.06,0.08\}$, respectively. Similarly, for MC=5, performance decreases from $14.7$ at $\beta=0$ to $14.4$, $13.7$, $13.6$, and $13.3$. Thus, the strongest TTT configuration corresponds to $\beta=0$, which exactly recovers QMAX.

\section{Compute Requirements, Limitations, and Broader Impact}

\textbf{Compute Requirements.}
All our experiments that were conducted on simulation were run on NVIDIA V100 GPUs including training, evaluation, and TTA (each experiment was run on a single GPU).
Regarding the real world experiments, we ran our TTA pipeline on the Unitree Go2 robot leveraging four CPU cores (2 for inference stream and 1 for adaptation and 1 for odometry).

\textbf{Limitations.}
Our study focuses on reward-free, critic-guided test-time adaptation for locomotion policies under deployment-time distribution shifts, and some limitations remain. 
First, while TEMPO is designed to be lightweight and on-device friendly, it still relies on access to a pretrained critic that is reasonably informative under shift; if the critic is severely miscalibrated or trained on a highly mismatched domain, adaptation quality may degrade. 
Second, TEMPO induces an additional, albeit small, memory and compute cost during inference.

\textbf{Broader Impact.}
This work aims to improve the robustness of robotic systems under deployment-time distribution shifts, reducing the need for manual recalibration and improving autonomy. However, adaptive systems may introduce risks if deployed without adequate testing and oversight, especially in safety-critical environments.
We emphasize that our method is intended to complement, not replace, careful system design, and we encourage responsible use in real-world settings.



\end{document}